\documentclass[11pt]{article}

\usepackage[final, nohyperref]{acl}

\usepackage[colorlinks, linkcolor=mydarkblue, 
urlcolor=mydarkblue, 
citecolor=mydarkblue]{hyperref}
\usepackage{xcolor}
\definecolor{mydarkblue}{HTML}{000099}
\definecolor{mydarkred}{rgb}{0.6,0,0}
\definecolor{myblue}{HTML}{268BD2}
\definecolor{mygreen}{HTML}{658354}

\usepackage{times}
\usepackage{latexsym}
\usepackage{textcomp}

\usepackage[T1]{fontenc}

\usepackage[utf8]{inputenc}

\usepackage{microtype}

\usepackage{inconsolata}

\usepackage{graphicx}

\usepackage{xspace}
\usepackage{tabularray}
\usepackage{tabularx}
\usepackage{multirow}
\usepackage{url}            
\usepackage{booktabs}       
\usepackage{amsfonts}       

\usepackage{amssymb}
\usepackage{bbm}
\usepackage[most]{tcolorbox}
\usepackage{multicol}
\usepackage{enumitem}
\usepackage{xurl}
\usepackage{tabularray}
\usepackage[htt]{hyphenat}
\usepackage[table]{xcolor}

\usepackage{tocloft}
\usepackage{titletoc}
\usepackage{etoolbox}
\usepackage{titlesec}

\usepackage{tocbibind}

\definecolor{goodcolor}{HTML}{2A9D8F}
\definecolor{badcolor}{HTML}{E76F51}

\newtcolorbox{researchquestion}{
  width=\linewidth,
  colback=goodcolor!10!white, 
  colframe=goodcolor!60!black,
  coltext=goodcolor!85!black,
  left=0.75mm,
  right=0.75mm,
  top=0.7mm,
  bottom=0.4mm,
  before upper=\textbf{Research Question: }\ignorespaces
}

\newtcolorbox{goodimplication}{
  width=\linewidth,
  colback=goodcolor!10!white, 
  colframe=goodcolor!60!black,
  coltext=goodcolor!85!black,
  left=0.75mm,
  right=0.75mm,
  top=0.7mm,
  bottom=0.4mm,
  before upper=\textbf{Good implication. }\ignorespaces
}

\newtcolorbox{badimplication}{
  width=\linewidth,
  colback=badcolor!10!white, 
  colframe=badcolor!60!black,
  coltext=badcolor!85!black,
  left=0.75mm,
  right=0.75mm,
  top=0.7mm,
  bottom=0.4mm,
  before upper=\textbf{Bad implication. }\ignorespaces
}

\newtcolorbox{prompt}{
  width=\linewidth,
  colback=gray!10!white, 
  colframe=gray!60!black,
  coltext=black,
  left=0.75mm,
  right=0.75mm,
  top=0.7mm,
  bottom=0.4mm,
  enhanced jigsaw,  
  enforce breakable,
}

\newcommand{\papertitle}{How Retrieved Context Shapes Internal Representations in RAG}
\title{\papertitle}

\newcommand{\eg}{{\it e.g.}\xspace}
\newcommand{\ie}{{\it i.e.}\xspace}

\ifx\definition\undefined
\newtheorem{definition}{Definition}[section]
\fi

\ifx\theorem\undefined
\newtheorem{theorem}{Theorem}[section]
\fi

\ifx\lemma\undefined
\newtheorem{lemma}{Lemma}[section]
\fi

\definecolor{mypurple}{HTML}{4C3C83}
\definecolor{myyellow}{HTML}{a98467}

\usepackage{amsmath,amsfonts,bm}

\def\eqref#1{equation~\ref{#1}}

\def\1{\bm{1}}

\def\vb{{\bm{b}}}

\def\vw{{\bm{w}}}
\def\vx{{\bm{x}}}

\DeclareMathAlphabet{\mathsfit}{\encodingdefault}{\sfdefault}{m}{sl}
\SetMathAlphabet{\mathsfit}{bold}{\encodingdefault}{\sfdefault}{bx}{n}

\author{Samuel Yeh \and Sharon Li \\
  Department of Computer Science, University of Wisconsin-Madison \\
  \texttt{\{samuelyeh, sharonli\}@cs.wisc.edu} \\}

\begin{document}
\maketitle
\begin{abstract}
Retrieval-augmented generation (RAG) enhances large language models (LLMs) by conditioning generation on retrieved external documents, but the effect of retrieved context is often non-trivial. In realistic retrieval settings, the retrieved document set often contains a mixture of documents that vary in relevance and usefulness. While prior work has largely examined these phenomena through output behavior, little is known about \textit{how retrieved context shapes the internal representations that mediate information integration in RAG}. In this work, we study RAG through the lens of latent representations. We systematically analyze how different types of retrieved documents affect the hidden states of LLMs, and how these internal representation shifts relate to downstream generation behavior. Across four question-answering datasets and three LLMs, we analyze internal representations under controlled single- and multi-document settings. Our results reveal how context relevancy and layer-wise processing influence internal representations, providing explanations of LLMs' output behaviors and insights for RAG system design.

\end{abstract}

\section{Introduction}

Retrieval-augmented generation (RAG) has become a widely adopted approach for enhancing large language models (LLMs) with external knowledge~\citep{10.1145/3637528.3671470, ram-etal-2023-context, izacard-grave-2021-leveraging, 10.5555/3495724.3496517}. 
By grounding generation in external evidence, RAG has been shown to improve factual accuracy, enhance coverage of long-tail knowledge, and enable dynamic knowledge updates without retraining the underlying model~\citep{10.5555/3524938.3525306, 10.1145/3626772.3657923, frisoni-etal-2022-bioreader, ji-etal-2023-rho, mallen-etal-2023-trust}.

However, the effect of retrieved context in RAG is not always straightforward. In realistic retrieval settings, the retrieved document set often contains a mixture of documents that vary in relevance and usefulness. While relevant documents can substantially improve performance, semantically similar but unhelpful documents can degrade generation quality~\citep{10.5555/3618408.3619699, fang-etal-2024-enhancing, wu-etal-2025-pandoras}. These observations question the reliability of RAG systems, calling for an in-depth understanding of how RAG truly works. 

\begin{figure*}
    \centering
    \includegraphics[ width=\textwidth]{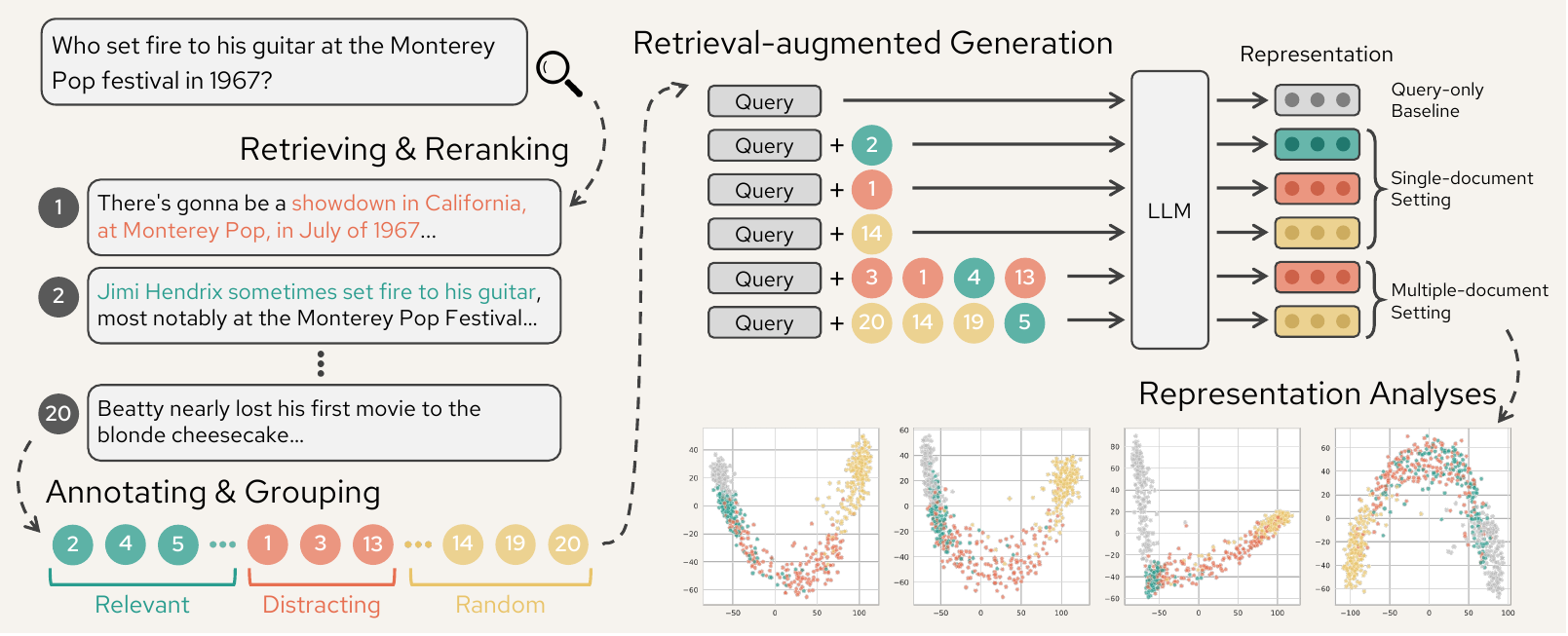} 
    \caption{\textbf{The overview of our analysis framework.} For each query, we retrieve and rerank a set of documents, and group them as relevant, distracting, and random. We then control the input context of RAG with different type(s) of documents and obtain the hidden representations for comparative analysis.}
    \label{fig:overview}
\end{figure*}

To fill the gap of understanding, prior work has largely focused on analyzing RAG at the level of \emph{output behavior}, studying how different retrieval strategies or context compositions affect final answers, such as accuracy~\citep{10.5555/3618408.3619699, wu-etal-2025-pandoras, vladika-matthes-2025-influence} and hallucination rates~\citep{joren2025sufficient, amiraz-etal-2025-distracting}. While such analyses provide important insights, they offer a limited understanding of how different types of retrieved documents influence the internal representations of LLMs.
Output-level observations alone cannot distinguish whether a change in output arises from effective evidence integration, suppression of parametric knowledge, or a deliberate model response such as uncertainty or refusal. A principled understanding of RAG, therefore, requires examining the internal representations that mediate interactions between retrieved information and a model’s parametric knowledge. {Along this line, \citet{wadhwa2024ragsrichparametersprobing} probed the LLM's internal representations to understand how LLMs behave when a relevant context is provided. However, it remains opaque how the internal representations evolve correspondingly in a more realistic setting, where the retrieved documents are mixtures of relevant information and noise, limiting both interpretability and informed system design.}

In this work, we study RAG through the lens of latent representation. We systematically analyze how retrieved documents shape the internal representations of LLMs and how internal representations relate to LLMs' performance, as illustrated in Figure~\ref{fig:overview}. Specifically, we examine the hidden states under controlled retrieval settings, varying the relevance of retrieved documents (relevant, distracting, and random), their combinations, and their interaction with query difficulty and model-internal knowledge. Across four question-answering datasets and three LLMs, our analyses reveal how context relevancy and layer-wise processing influence internal representations. 

Our results reveal consistent and previously underexplored patterns. 
Relevant documents often leave representations largely unchanged, acting primarily to reinforce existing parametric knowledge rather than introducing decisive new information. Layer-wise analyses further show that later layers increasingly emphasize parametric knowledge, limiting the influence of retrieved evidence. In contrast, random documents trigger large representation shifts that are closely tied to abstention behavior, indicating that models internally recognize uninformative context and transition into a refusal mode. We also find that in multi-document settings, a single relevant document can anchor internal representations and suppress the influence of additional noise.

Beyond explanation, our findings also yield practical insights for RAG system design, highlighting when noisy context can be safely tolerated and when retrieved evidence fails to meaningfully influence generation. Together, our work offers a representation-level perspective on RAG that complements prior output-focused analyses and advances understanding of how LLMs internally utilize retrieved context. Our key contributions are summarized as follows:

\begin{enumerate}[leftmargin=*, itemsep=1pt]
    \item We propose a rigorous framework for analyzing internal representations in RAG, enabling controlled and systematic study of how retrieved context is processed by LLMs.
    \item We uncover new representation-level patterns and derive practical implications across different document relevance types, query difficulties, and model settings.
    \item We connect these representation-level findings to observable LLM behaviors, providing mechanistic explanations for RAG phenomena and actionable insights for RAG system design.
\end{enumerate}

\section{Related Work}
\paragraph{Reliable retrieval-augmented generation.}

Recent work in RAG has increasingly focused on the reliability issue. Among them, empirical studies have shown that retrieved context often contains heterogeneous signals, including relevant evidence, semantically similar but misleading passages, and irrelevant noise, all of which can substantially influence model behavior~\citep{10.1145/3626772.3657834, wu-etal-2025-pandoras, xu-etal-2024-knowledge-conflicts, yang2025quantifyingrobustnessretrievalaugmentedlanguage, 10.5555/3618408.3619699}. Prior work has examined how LLMs respond to knowledge conflicts between retrieved documents and internal knowledge, revealing behaviors such as selective reliance on retrieval, stubborn adherence to parametric knowledge, or instability under conflicting evidence~\citep{xie2024adaptive, wadhwa2024ragsrichparametersprobing}. Other studies have highlighted the effects of noisy or irrelevant documents on generation quality, including distraction~\citep{amiraz-etal-2025-distracting}, hallucination~\citep{joren2025sufficient}, and shown that LLMs are sensitive to document ordering~\citep{liu-etal-2024-lost, cuconasu-etal-2025-rag} and context length~\citep{levy-etal-2024-task}. Recent benchmarks and evaluation frameworks further explored RAG performance in long-context and long-form generation settings, emphasizing challenges related to context utilization, retrieval coverage, and robustness~\citep{ju-etal-2025-controlled, qi-etal-2024-long2rag}.

Motivated by these findings, a range of methods have been proposed to improve RAG reliability, including selective context filtering~\citep{he-etal-2025-select, xu2024recomp, zhu-etal-2024-information, deng2025influence}, document reranking~\citep{wang-etal-2025-infogain}, and knowledge-aware decoding~\citep{10.1145/3767695.3769515, wang2025retrievalaugmented, sun2025lfdlayerfuseddecoding, xiang2024certifiably}. While effective at improving output-level metrics such as accuracy and hallucination rates, these approaches leave open how retrieved documents influence internal model representations.

\paragraph{Hidden representation of LLMs.}

A long line of research has aimed to study the internal representations of LLMs and shown that these representations encode linguistic, semantic, and task-relevant information~\citep{liu-etal-2019-linguistic, tenney-etal-2019-bert, voita-etal-2019-bottom, jin-etal-2025-exploring, gurnee2024language, 10.24963/ijcai.2025/566}.
Prior work has also shown that internal states evolve in structured ways across layers, reflecting the progression from lexical processing to higher-level semantic and decision-related representations~\citep{skean2025layer}. Recent studies further reveal that internal states often contain signals of uncertainty, hallucination, or knowledge conflict even when outputs appear confident~\citep{azaria-mitchell-2023-internal, chen2024inside, du2024haloscope}.

{In RAG settings, \citet{wadhwa2024ragsrichparametersprobing} probed internal representations and found that LLMs tend to have a strong bias towards utilizing only the context information to answer the question when a relevant document is provided.} Other recent work has begun to exploit internal representations to assess the faithfulness of generation. 
However, existing studies primarily used hidden representations as tools for downstream tasks, such as hallucination and knowledge conflict detection~\citep{yeh2025lumina, zhao2024analysing}, without systematically analyzing how different types of retrieved context shape internal states in the first place. In contrast, our work adopts representations as tools for analysis, understanding how retrieved context is internally processed in RAG, and how these internal dynamics relate to downstream generation behavior.

\section{Definition and Problem Statement}

\begin{definition}[{Retrieval-Augmented Generation.}]
    Given a query $q$, a retriever $r:q\mapsto S_q$ fetches a set of $N$ documents $S_q=\{d_1,
    \dots,d_N\}\subset\mathcal{D}$ from a database $\mathcal{D}$. Retrieval-augmented generation (RAG) conditions an LLM $p_\theta$ on both the query and the retrieved documents to generate an answer:
    \begin{align}
        \hat{y}\sim p_\theta(Y|I, q, S_q),
    \end{align}
    where $I$ denotes the instruction prompt.
\end{definition}

In practice, $d_i$ can be viewed as a document sampled from a mixture of three distributions: \emph{relevant} ($\mathbb{P}_{\text{rel}}$), \emph{distracting} ($\mathbb{P}_{\text{dist}}$), and \emph{random} ($\mathbb{P}_{\text{rand}}$) documents, \ie,
\begin{align}
d_i \sim \alpha_1 \mathbb{P}_{\text{rel}} + \alpha_2 \mathbb{P}_{\text{dist}} + \alpha_3 \mathbb{P}_{\text{rand}},
\end{align}
where $\alpha_1, \alpha_2, \alpha_3 > 0$ and $\alpha_1 + \alpha_2 + \alpha_3 = 1$. We define these document types as follows:
\begin{itemize}[leftmargin=*, itemsep=1pt]
    \item \textbf{Relevant.} A document $d_i$ is relevant to query $q$ if it contains the ground-truth answer $y$ or provides partial information that directly supports  $y$.
    \item \textbf{Distracting.} A document $d_i$ is distracting if it exhibits high semantic similarity to $q$ but does not contain information that supports deriving $y$, and may potentially mislead the model.
    \item \textbf{Random.} A document $d_i$ is random if it has low semantic similarity to $q$ and does not contain information helpful for deriving $y$.
\end{itemize}
An example of each type of document is presented in Appendix~\ref{ap:examples}.

Let $h^{q,S_q} \in \mathbb{R}^{L \times D}$ denote the hidden states at the last prompt token across all $L$ transformer layers, given query $q$ and document set $S_q$, where $D$ is the hidden dimension. In this work, we study the relationship between the retrieved document set $S_q$, the resulting internal representations $h^{q,S_q}$, and the generated answer $\hat{y}$. Specifically, we aim to address the following research questions:

\begin{researchquestion}
   \emph{How do different types of retrieved documents influence the internal representations that govern generation in RAG?}
\end{researchquestion}

\section{Analysis Setup}
\paragraph{Overview.}

We design a controlled experimental setup to analyze how retrieved documents of varying relevance affect LLM internal representations and downstream generation. By fixing the RAG pipeline and systematically varying document relevance and context composition, we isolate representation changes attributable to retrieved evidence and relate them to downstream generation behavior.

\subsection{Settings of RAG and Data}\label{sec:rag_setting}

\paragraph{Datasets.}
We study the impact of RAG on four representative datasets, including Trivia QA~\citep{joshi-etal-2017-triviaqa}, NQ~\citep{kwiatkowski-etal-2019-natural}, Pop QA~\citep{mallen-etal-2023-trust}, and Strategy QA~\citep{geva-etal-2021-aristotle}. The details of each dataset are provided in Appendix~\ref{ap:dataset}.

\paragraph{Models.}
We conduct experiments with LLMs of varying sizes, including \mbox{Gemma3-27B}~\citep{gemmateam2025gemma3technicalreport}, \mbox{Llama4-17B}~\citep{llama4}, and \mbox{Qwen3-Next-80B}~\citep{yang2025qwen3technicalreport}.

\paragraph{Retrieval database \& algorithm.}
We employ MassiveDS~\citep{NEURIPS2024_a5d8aba2} as the retrieval database, which comprises 1.4 trillion tokens.  We use Contriever~\citep{izacard2022unsupervised} to retrieve the top 20 documents for each query. We discuss the detailed retrieval setting in Appendix~\ref{ap:retrieval}. 

\paragraph{Query difficulty categorization.}

To account for the impact of a model's internal knowledge on representations, we categorize queries for each LLM based on whether it can correctly answer them \emph{without} retrieved documents. Specifically, for each query $q$, we prompt each model $p_\theta$ using the query alone and evaluate the generated answer $\hat{y}$ against the ground truth $y$ using Qwen3-Next-80B as a judge (see Appendix~\ref{ap:prompt} for the prompt design and human verification).\footnote{{We use Qwen3-Next-80B for both response and ground truth label generation due to a budget constraint. To verify reliability, we manually re-annotated 50 randomly sampled instances and found only 1 discrepancy with the LLM judgments, indicating strong agreement.}} Queries that are correctly answered without retrieval are labeled as {\emph{easy}} for that model, while the others are labeled as {\emph{hard}}. The statistics of query difficulty of each dataset and LLM are shown in Appendix~\ref{ap:query_difficulty}.

\paragraph{Document set formulation.}
For each retrieved document $d_i \in S_q$, we classify whether $d_i$ is relevant or distracting with respect to query $q$ and ground-truth answer $y$. We perform this classification by prompting GPT-5 (see Appendix~\ref{ap:prompt} for the prompt design and human verification). Documents not classified as relevant or distracting are treated as neither category. Based on this classification, we construct three document sets for each query $q$:
\begin{enumerate}[itemsep=1pt]
    \item $\mathcal{S}_{q}^{\text{rel}} := \{ d_i \in S_q \mid d_i \text{ is relevant} \}$,
    \item $\mathcal{S}_{q}^{\text{dist}} := \{ d_i \in S_q \mid d_i \text{ is distracting} \}$,
    \item $\mathcal{S}_{q}^{\text{rand}} := S_{q'},$ where $q'$ is a randomly sampled query from the same dataset.
\end{enumerate}
Examples of each document type are shown in Appendix~\ref{ap:examples}.
These document sets serve as reusable building blocks for the representation analysis settings described next.

\begin{figure*}[!t]
    \centering
    \includegraphics[width=\textwidth]{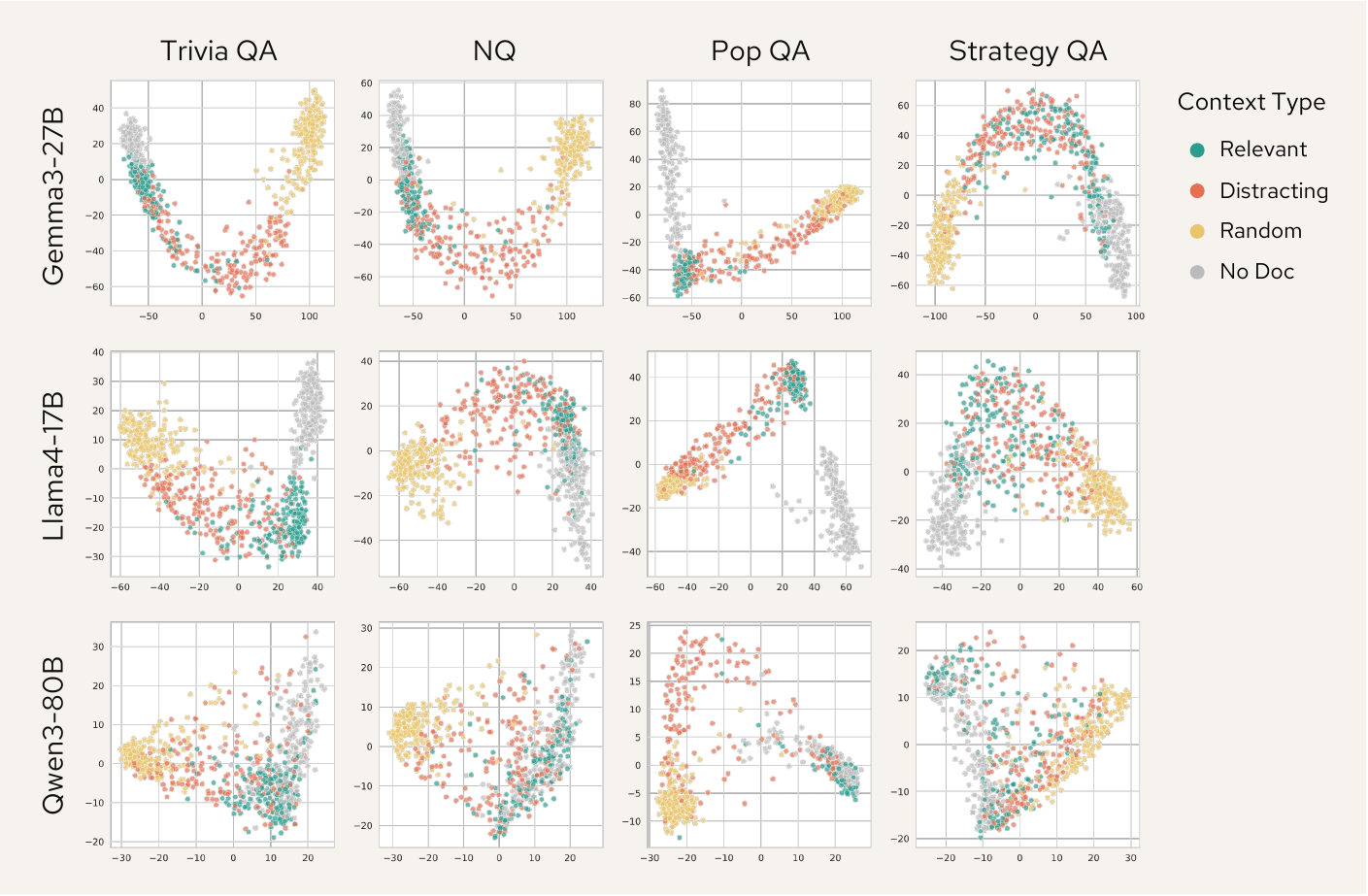} 
    \caption{\textbf{Representations of prompts paired with semantically similar documents remain close to the no-document baseline, whereas semantically dissimilar documents cause the representations to drift away.} We apply PCA on the last prompt token representations across different document types and plot them in 2D.}
    \label{fig:single_document}
\end{figure*}

\subsection{Settings of Representation Analysis}\label{sec:representation_setting}

\paragraph{Single-document setting.}

In this setting, we analyze the effect of individual documents by constructing prompts that include one document per query: (\textit{i}) a relevant document from $\mathcal{S}_{q}^{\text{rel}}$, (\textit{ii}) a distracting document from $\mathcal{S}_{q}^{\text{dist}}$, (\textit{iii}) a random document from $\mathcal{S}_{q}^{\text{rand}}$, or (\textit{iv}) no document (query-only baseline). This setting isolates the impact of document relevance on internal representations.

\paragraph{Multiple-document setting.}

In this setting, we consider a realistic RAG scenario with multiple documents per query. Prompts contain four documents under two conditions: (\textit{i}) one relevant document paired with three distracting documents, or (\textit{ii}) one relevant document paired with three random documents, along with a relevant-only baseline. Documents are \emph{randomly shuffled} to reduce positional bias. This setting examines how relevant evidence is represented among competing contexts.

\section{Analysis Results}\label{sec:result}

\subsection{Effect of Context Relevancy}

Relevance of retrieval documents plays an important role in RAG systems: context can only help generation when it contains relevant information w.r.t. the ground truth. Beyond retrieval quality, LLMs must also appropriately utilize the provided context, \eg, integrating informative content while ignoring irrelevant or noisy documents. 

Intuitively, we may expect relevant documents to introduce information beyond the model's parametric knowledge and thus shift representations away from the query-only state, especially for queries the model cannot answer without retrieval. In contrast, random documents are expected to carry little useful signal and thus induce minimal representation change. To examine how context relevance affects internal representations, we apply principal component analysis (PCA) to $h^{q,S_q}_{-1}$, \ie, the final layer representations of the last prompt token, under different context types and visualize them in two dimensions. Because the hidden representation of the final prompt token directly conditions the output token distribution, differences in how retrieved documents are processed should be reflected in this representation. Beyond visualizations, we further validate our findings in Appendix~\ref{ap:separability} via quantitative analyses of representation separability. 

\paragraph{Observation 1: Random documents induce large representation drift.}

Figure~\ref{fig:single_document} shows the PCA visualization under the single-document setting. Contrary to intuition, random documents induce substantial representation drift from the query-only baseline, often larger than that caused by distracting or even relevant documents.

\begin{figure}[!t]
    \centering
    \includegraphics[width=\linewidth]{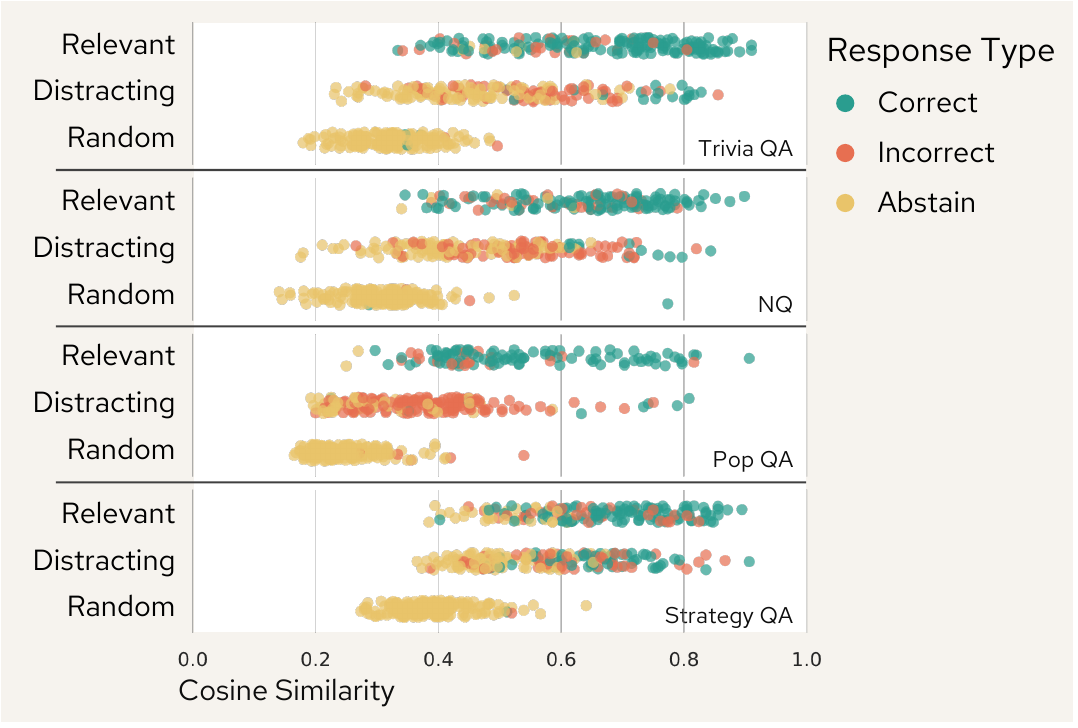} 
    \caption{\textbf{LLMs are more likely to abstain when context induces large representation shifts.} For each context type, we compute the cosine similarity between the representations of with-context prompts and query, and categorize responses as correct, incorrect, or abstain. We show the result of Gemma3-27B. For other models, see Figure~\ref{fig:sim_vs_response_full}.}
    \label{fig:sim_vs_response}
\end{figure}

\begin{figure}[!t]
    \centering
    \includegraphics[ width=\linewidth]{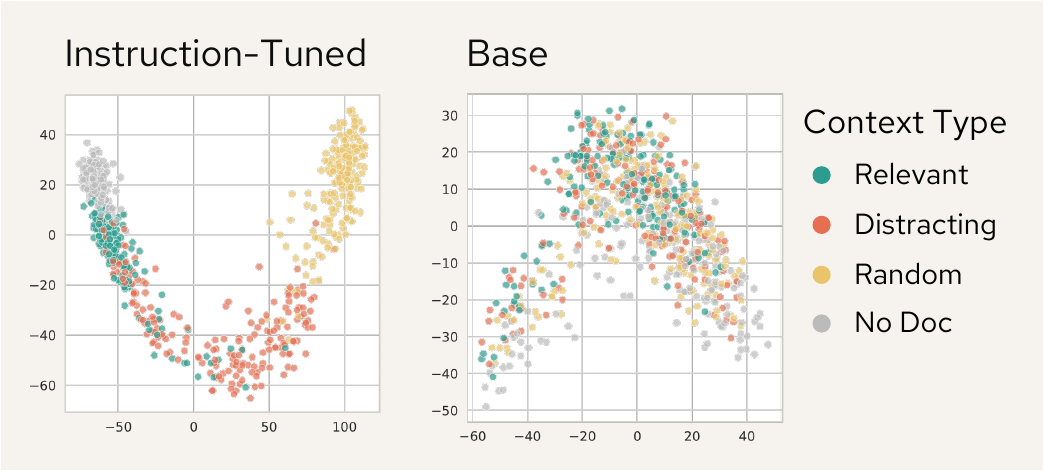} 
    \caption{\textbf{Base LLMs do not have representation drifts across different context types.} We apply PCA on representations of instruction-tuned models and base models. We show the result of Gemma3-27B on Trivia QA. For other models and datasets, see Figure~\ref{fig:base_vs_it_full}.}
    \label{fig:base_vs_it}
\end{figure}

This representation drift is strongly linked to output behavior.  Figure~\ref{fig:sim_vs_response} plots response categories against the cosine similarity between with- and without-context representations. The result shows that models are significantly more likely to abstain when the with-context representations are highly dissimilar from the without-context representations. This suggests that LLMs internally recognize the lack of useful information in random context and shift toward a refusal mode, which manifests as large representation drift.
We further investigate the origin of this behavior by repeating the analysis with base models. As shown in Figure~\ref{fig:base_vs_it}, representation drift from random documents largely vanishes, and Table~\ref{tb:base_vs_it} shows that base models abstain far less frequently ($<20\%$) than instruction-tuned models ($>60\%$). This suggests that while abstention exists in base models, it is substantially amplified by instruction tuning.

This behavior has mixed practical implications. On the one hand, abstention under clearly uninformative context reflects an internal mechanism for avoiding unsupported answers.
On the other hand, representation drift from random documents occurs for both easy and hard queries, causing instruction-tuned models to abstain even when they could answer using parametric knowledge alone. As Table~\ref{tb:base_vs_it} shows, base models retain strong performance on easy queries under random context, whereas this ability largely disappears after instruction tuning. This behavior undermines RAG usability: an ideal model should signal missing evidence while still answering when sufficient internal knowledge is available.

\begin{goodimplication}
    LLMs internally detect uninformative context and appropriately abstain from answering, {especially after instruction tuning and when their parametric knowledge does not cover the answer, ensuring the reliability of LLMs under noisy scenarios.} 
\end{goodimplication}

\begin{badimplication}
    Instruction tuning can suppress reliance on parametric knowledge in the presence of irrelevant context, even when correct answers are available without retrieval, {weakening the usability of RAG systems.}
\end{badimplication}

\paragraph{Observation 2: Relevant documents largely preserve internal representations.}

Figure~\ref{fig:single_document} also shows that relevant documents induce relatively small representation shifts compared to the query-only condition. For easy queries, this behavior is expected as relevant documents typically align with models' parametric knowledge and therefore do not push representations toward a different region of the latent space. Consistent with this interpretation, we find that responses generated with relevant documents usually achieve significantly higher log-likelihood than query-only responses ($p<0.001$), indicating increased model confidence (see Appendix~\ref{ap:uncertainty}). In this regime, retrieved evidence primarily acts as a confirmation signal.

\begin{table}[!t]
    \centering
    \small
    \resizebox{\linewidth}{!}{
    \begin{tabular}{lcccccc}
        \toprule
         & \multicolumn{3}{c}{Easy} & \multicolumn{3}{c}{Hard}\\
         \cmidrule(lr){2-4}
         \cmidrule(lr){5-7}
         Context & Cor & Inc & Abs & Cor & Inc & Abs\\
        \midrule

        \rowcolor{gray!15}\multicolumn{7}{c}{Base}\\
        \midrule
        Relevant & 92.4 & 4.2 & 3.4 & 52.5 & 35.6 & 11.9\\
        Distracting & 79.5 & 15.8 & 4.7 & 14.8 & 72.4 & 12.8\\
        Random & 89.4 & 6.8 & 3.8 & 15.6 & 65.4 & 19.0\\

        \midrule
        \midrule

        \rowcolor{gray!15}\multicolumn{7}{c}{Instruction-tuned}\\
        \midrule
        Relevant & 90.4 & 6.5 & 3.1 & 65.2 & 27.8 & 7.0\\
        Distracting & 8.5 & 29.7 & 61.8 & 0.7 & 25.1 & 74.2\\
        Random & 1.7 & 0.7 & 97.6 & 0 & 1.9 & 98.1\\
        
        \bottomrule
    \end{tabular}
    }
    \caption{
    \textbf{Instruction-tuned LLMs tend to abstain when the retrieval document is distracting or random, even if they can answer with the query alone.} We report the percentage of correct (Cor), incorrect (Inc), and abstain (Abs) responses for both base and instruction-tuned LLMs. We show the result of Gemma3-27B on Trivia QA. For other models and datasets, see Table~\ref{tb:base_vs_it_full}.
    }
    \label{tb:base_vs_it}
\end{table}

In contrast, for the hard queries, the consistently small representation drifts indicate that relevant documents often fail to provide a sufficiently strong signal to meaningfully alter internal representations. Table~\ref{tb:base_vs_it} shows that for hard queries, 35.6\% of responses generated by base LLMs remain incorrect even when relevant documents are provided. In some cases, instruction-tuned LLMs exhibit an even higher error rate with relevant documents than with distracting documents. This indicates that when parametric knowledge is insufficient, relevant documents are not always effectively integrated and may introduce unresolved competition between weak parametric knowledge and retrieved evidence. These findings reveal the following practical implications of the current RAG system:

\begin{goodimplication}
    Relevant documents {primarily act as a confirmation signal to reinforce parametric knowledge}, increasing confidence and reliability for easy queries. 
\end{goodimplication}

\begin{badimplication}
    RAG has a limited impact on hard queries as relevant documents often fail to sufficiently influence internal representations when parametric knowledge is lacking. {Unfortunately, this is exactly when relevant documents are really needed.}
\end{badimplication}

\paragraph{Observation 3: A single relevant document stabilizes representations in multi-document settings}

\begin{figure}[!t]
    \centering
    \includegraphics[width=\linewidth]{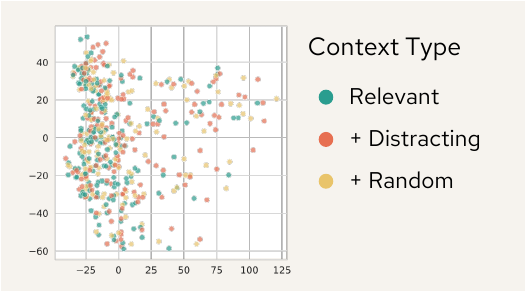} 
    \caption{\textbf{Representations remain similar when a relevant document is present, regardless of other context.} We perform PCA on the last prompt token representations for multiple-document contexts that include one relevant document, and plot them alongside representations with only a relevant document in 2D. We show the result of Gemma3-27B on Trivia QA. For other models and datasets, see Figure~\ref{fig:multiple_document_full}.}
    \label{fig:multiple_document}
\end{figure}

We next examine the multi-document setting, where relevant documents are presented alongside distracting or random documents. Figure~\ref{fig:multiple_document} shows that when at least one relevant document is included, the resulting representations remain close to those obtained with a relevant-only context. Unlike the single-document case (Figure~\ref{fig:single_document}), where distracting or random documents alone cause large representation drift, the presence of a relevant document anchors the model’s internal state despite additional noise.

This stability is mirrored in output behavior. As shown in Table~\ref{tb:accuracy}, accuracy is largely preserved and sometimes improved whenever a relevant document is present, regardless of additional distracting or random context, comparing to the distracting or random only baselines. Together, these results suggest that LLMs can selectively attend to informative evidence and suppress irrelevant signals when reliable grounding is available.

\begin{goodimplication}
    LLMs effectively prioritize relevant documents, yielding robust internal representations and generation even in the presence of noisy context.
\end{goodimplication}

\subsection{Effect of Layer-wise Process}

Beyond the final-layer representations, we investigate how internal representations evolve across layers. Understanding this layer-wise process is critical for characterizing when and how retrieved context begins to influence model behavior.

\paragraph{Observation 4: LLMs first distinguish between prompts with random documents from others.}

\begin{table}[!t]
    \centering
    \small
    \resizebox{\linewidth}{!}{
    \begin{tabular}{lcccccccc}
        \toprule
         & \multicolumn{2}{c}{Trivia QA} & \multicolumn{2}{c}{NQ} & \multicolumn{2}{c}{Pop QA} & \multicolumn{2}{c}{Strategy QA}\\
         \cmidrule(lr){2-3}
         \cmidrule(lr){4-5}
         \cmidrule(lr){6-7}
         \cmidrule(lr){8-9}
         Context & Easy & Hard & Easy & Hard & Easy & Hard & Easy & Hard\\
        \midrule

        \rowcolor{gray!15}\multicolumn{9}{c}{Gemma3-27B}\\
        \midrule
        Relevant & 90.4 & 65.2 & 79.8 & 62.1 & 91.0 & 70.5 & 72.7 & 44.3\\
        \rowcolor{goodcolor!30} + Distracting & 82.6 & 57.1 & 73.7 & 47.8 & 84.6 & 58.0 & 64.0 & 27.9\\
        \rowcolor{goodcolor!30} + Random & 87.7 & 60.2 & 79.7 & 48.6 & 84.0 & 69.8 & 58.3 & 30.8\\
        \midrule
        Distracting & 8.4 & 0.7 & 8.0 & 0.6 & 3.6 & 0.6 & 22.7 & 8.6\\
        Random & 1.7 & 0 & 2.2 & 0.4 & 0.2 & 0.1 & 1.2 & 0\\
        
        \midrule
        \midrule

        \rowcolor{gray!15}\multicolumn{9}{c}{Llama4-17B}\\
        \midrule
        Relevant & 89.8 & 62.8 & 85.9 & 64.0 & 92.1 & 79.8 & 76.3 & 35.3\\
        \rowcolor{goodcolor!30} + Distracting & 84.3 & 48.4 & 80.4 & 47.7 & 89.8 & 67.9 & 71.5 & 27.1\\
        \rowcolor{goodcolor!30} + Random & 90.6 & 55.5 & 86.9 & 65.1 & 91.7 & 83.7 & 73.2 & 34.3\\
        \midrule
        Distracting & 34.7 & 11.4 & 33.5 & 5.1 & 2.5 & 0.5 & 38.5 & 13.2\\
        Random & 38.8 & 8.7 & 34.5 & 4.0 & 0.4 & 0 & 13.9 & 13.2\\
        \midrule
        \midrule

        \rowcolor{gray!15}\multicolumn{9}{c}{Qwen3-Next-80B}\\
        \midrule
        Relevant & 93.8 & 64.5 & 88.9 & 66.7 & 94.6 & 81.0 & 84.9 & 44.0\\
        \rowcolor{goodcolor!30} + Distracting & 85.6 & 55.3 & 88.0 & 51.2 & 86.3 & 57.1 & 83.7 & 30.7\\
        \rowcolor{goodcolor!30} + Random & 93.6 & 66.7 & 90.6 & 62.8 & 92.2 & 93.8 & 86.1 & 39.1\\
        \midrule
        Distracting & 48.8 & 7.5 & 53.3 & 6.1 & 3.6 & 0.4 & 63.1 & 10.5\\
        Random & 33.9 & 2.8 & 22.3 & 1.2 & 0.5 & 0 & 46.4 & 7.2\\
        
        \bottomrule
    \end{tabular}
    }
    \caption{
    \textbf{Performance is preserved or even improved if at least one relevant document is presented in the input context.} We report the percentage of correct responses for each LLM and dataset across different context types. 
    }
    \label{tb:accuracy}
\end{table}

\begin{figure*}[!t]
    \centering
    \includegraphics[width=\textwidth]{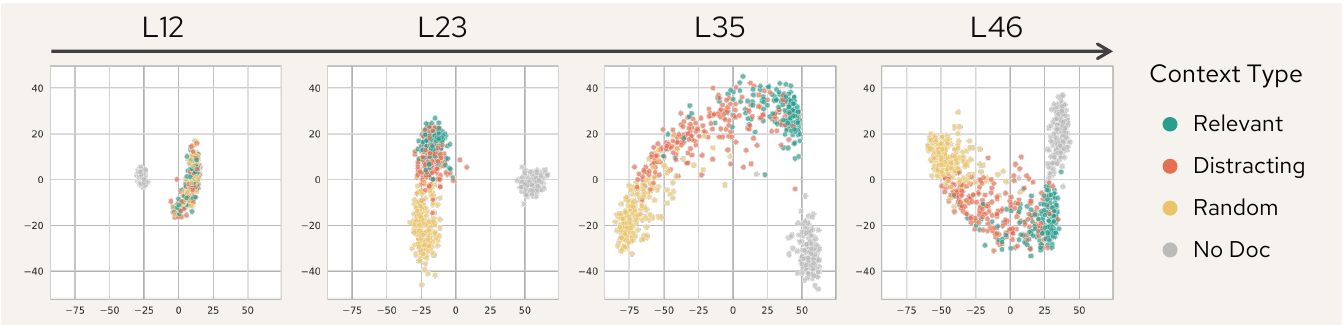} 
    \caption{\textbf{Representations with random documents are separated from others in earlier layers, and representations with relevant documents are drawn close to no-document representations in later layers.} We perform PCA on the last prompt token representations across different layers and document types and plot them in 2D. We show the result of Llama4-17B on Trivia QA. For other models and datasets, see Figure~\ref{fig:layer_gemma3_27B} to \ref{fig:layer_qwen3_80B}.}
    \label{fig:layer}
\end{figure*}

Figure~\ref{fig:layer} shows that in the early layers (L12), representations corresponding to prompts with relevant, distracting, and random documents largely overlap. After a certain layer (L23), representations associated with random documents become distinguishable from the others, forming a separate cluster. This observation suggests that coarse semantic mismatches between the query and the input context are relatively easy for LLMs to identify and can be detected early in the processing pipeline.

In contrast, representations corresponding to relevant and distracting documents remain highly intertwined until much later layers (L35). Discriminating between these two context types likely requires higher-level semantic reasoning and integration with the model's parametric knowledge. While partial separation between relevant and distracting contexts emerges in deeper layers, the representations remain not fully separable even at the final layer. This suggests that current LLMs have limited capacity to reliably distinguish hard-negative noise from genuinely informative evidence.

\begin{goodimplication}
    LLMs can detect clearly uninformative or random context early in the internal processing pipeline.
\end{goodimplication}

\begin{badimplication}
    Differentiating relevant documents from distracting ones requires deep processing and remains imperfect, limiting robustness under noisy retrieval.
\end{badimplication}

\paragraph{Observation 5: Later layers close the gap between no-document and relevant-document representations.}

Another phenomenon we observed through Figure~\ref{fig:layer} is that the representations without documents stay far away from representations with documents in earlier layers. However, after the middle layer (L35), the representations with relevant documents gradually move toward the no-document representations and become close in later layers (L46). This convergence suggests that later layers play a dominant role in reconciling retrieved evidence with the model's parametric knowledge. In Figure~\ref{fig:layer_gemma3_27B-base} and \ref{fig:layer_llama4_17B-base}, we further show that the dominance of parametric knowledge in later layers also exists in base models, suggesting that it is more like a property of the transformer decoder than an artifact of instruction tuning.

This layer-wise pattern explains the overlap observed in Observation 2. While the information introduced by relevant documents is retained in intermediate-layer representations, the final layers progressively emphasize the model's internal knowledge, reducing the influence of retrieved context. Although this behavior helps remove noise and stabilize generation when retrieved information is consistent with parametric knowledge, it also limits the impact of RAG for hard queries, where parametric knowledge is insufficient, and greater reliance on external evidence would be necessary.

\begin{goodimplication}
Later layers integrate retrieved evidence with parametric knowledge, reducing residual noise and stabilizing generation.
\end{goodimplication}

\begin{badimplication}
The increasing dominance of parametric knowledge in later layers weakens the influence of retrieved evidence, limiting RAG effectiveness on hard queries.
\end{badimplication}

\section{Discussion}

\paragraph{Representation analyses provide insights on construction of input context.}

Our representation analyses yield practical guidance for RAG design. 
For example, Observation 3 shows that when at least one relevant document is present, LLMs maintain stable internal representations and are robust to additional distracting or random context. This suggests that increasing retrieval breadth to improve recall can be beneficial, as long as there is a reasonable chance of retrieving at least one relevant document. We validate this implication by providing models with the full set of 20 retrieved documents without filtering. As shown in Table~\ref{tb:all_doc}, this unfiltered setting achieves performance comparable to using only a relevant document, indicating that LLMs can internally suppress noise when reliable evidence is available, reducing the need for aggressive document filtering in some RAG setups.

\begin{table}[!t]
    \centering
    \small
    \resizebox{\linewidth}{!}{
    \begin{tabular}{lcccccccc}
        \toprule
         & \multicolumn{2}{c}{Trivia QA} & \multicolumn{2}{c}{NQ} & \multicolumn{2}{c}{Pop QA} & \multicolumn{2}{c}{Strategy QA}\\
         \cmidrule(lr){2-3}
         \cmidrule(lr){4-5}
         \cmidrule(lr){6-7}
         \cmidrule(lr){8-9}
         Context & Easy & Hard & Easy & Hard & Easy & Hard & Easy & Hard\\
        \midrule

        \rowcolor{gray!15}\multicolumn{9}{c}{Gemma3-27B}\\
        \midrule
        Relevant Only & 90.4 & 65.2 & 79.8 & 62.1 & 91.0 & 70.5 & 72.7 & 44.3\\
        20 Documents & 90.3 & 61.5 & 89.2 & 53.5 & 85.1 & 69.1 & 71.5 & 24.2\\
        
        \midrule
        \midrule

        \rowcolor{gray!15}\multicolumn{9}{c}{Llama4-17B}\\
        \midrule
        Relevant Only & 89.8 & 62.8 & 85.9 & 64.0 & 92.1 & 79.8 & 76.3 & 35.3\\
        20 Documents & 94.7 & 43.8 & 91.4 & 57.1 & 87.3 & 65.4 & 72.6 & 30.4\\

        \midrule
        \midrule

        \rowcolor{gray!15}\multicolumn{9}{c}{Qwen3-Next-80B}\\
        \midrule
        Relevant Only & 93.8 & 64.5 & 88.9 & 66.7 & 94.6 & 81.0 & 84.9 & 44.0\\
        20 Documents & 92.4 & 68.0 & 95.5 & 52.4 & 90.8 & 58.5 & 79.5 & 47.9\\
        
        \bottomrule
    \end{tabular}
    }
    \caption{
    \textbf{Performance is preserved when inputting 20 documents without filtering according to relevancy.} We report the percentage of correct responses for each LLM and dataset across different context types. 
    }
    \label{tb:all_doc}
\end{table}

\paragraph{Representation analyses point toward principled interventions for hard queries.}

{Beyond diagnosis, our findings suggest concrete directions for improving RAG on hard queries. The layer-wise analysis reveals that later layers increasingly favor parametric knowledge, weakening the influence of retrieved context. This points toward potential interventions such as (i) modifying decoding or fusion strategies to amplify retrieval signals in later layers, (ii) architectural changes that allocate dedicated retrieval-sensitive subspaces, or (iii) training objectives that explicitly reward representation shifts toward external evidence when parametric knowledge is insufficient.}

\paragraph{Conclusion.}
In this work, we study RAG from a representation-level perspective, analyzing how LLMs internally process retrieved context. By examining last prompt token representations across document relevance types, query difficulty levels, and transformer layers, we complement prior output-focused analyses of RAG. We find that semantically dissimilar context induces large representation drift associated with abstention, while relevant documents largely preserve representation geometry and primarily act as confirmation signals, especially for easy queries. In multi-document settings, a single relevant document stabilizes internal representations and mitigates the impact of additional noise. Layer-wise analyses further show that coarse mismatches are detected early, whereas later layers increasingly favor parametric knowledge, limiting the influence of retrieved evidence on hard queries. Together, our findings provide mechanistic explanations for phenomena such as the distracting effect and confirmation bias, and offer insights for designing more reliable RAG systems.

\section*{Ethical Consideration}

This work focuses on analyzing the internal representations of LLMs in RAG systems. Our study is empirical and diagnostic in nature, aiming to improve understanding of how retrieved context influences model behavior rather than proposing new systems. All datasets used in our experiments are publicly available datasets for question answering and do not contain personally identifiable or sensitive information. In addition, we do not collect new data or involve human subjects.

\section*{Limitation}

{Our analysis focuses primarily on the last prompt token representations and their evolution across layers because the last prompt token directly conditions the output distribution and therefore provides a clean, comparable representation across different retrieval settings. However, the single token representation does not capture token-level dynamics within the context or during response generation.} Extending representation analysis to earlier prompt tokens or to response tokens could provide a more fine-grained understanding of RAG behavior.

\section*{Disclosure of LLM Usage}

In this work, we use LLMs-as-a-Judge to classify responses and the relevancy of each retrieved document, as many studies have shown that automated metrics such as F1-score and exact match are less reliable than LLMs-as-a-Judge~\citep{li2025lara, joren2025sufficient}. To ensure transparency and reliability, we provide the prompt we used and the result of human verification in Appendix~\ref{ap:prompt}. For writing, we only use LLMs to check grammar and paraphrase unnatural sentences in order to enhance readability.

\section*{Acknowledgement}

The authors would like to thank Hyeong Kyu Choi and Jiatong Li for their valuable feedback on the draft. This work is supported by AFOSR Young Investigator Program under award number FA9550-23-1-0184, National Science Foundation under awards IIS-2237037 and IIS-2331669, Office of Naval Research under grant number N00014-23-1-2643, Schmidt Sciences Foundation, Open Philanthropy, and Alfred P. Sloan Fellowship.

\bibliography{acl}

\newpage

\appendix
\onecolumn
\textsc{\huge {Appendix}}

\addcontentsline{toc}{section}{Appendix}

\startcontents[appendix]

\vspace{1.5em}
\textsc{\Large Contents}

\begingroup
  \setcounter{tocdepth}{2}
  \printcontents[appendix]{l}{1}{}
\endgroup
\twocolumn

\begin{table*}[!t]
    \centering
    \small
    \begin{tabular}{lcccccccc}
        \toprule
         & \multicolumn{2}{c}{Trivia QA} & \multicolumn{2}{c}{NQ} & \multicolumn{2}{c}{Pop QA} & \multicolumn{2}{c}{Strategy QA}\\
         \cmidrule(lr){2-3}
         \cmidrule(lr){4-5}
         \cmidrule(lr){6-7}
         \cmidrule(lr){8-9}
         LLM & Easy & Hard & Easy & Hard & Easy & Hard & Easy & Hard\\
        \midrule

        Gemma3-27B & 150 & 50 & 117 & 83 & 66 & 134 & 155 & 45 \\
        Llama4-17B & 158 & 42 & 118 & 82 & 93 & 107 & 140 & 60\\
        Qwen3-80B & 159 & 41 & 121 & 79 & 152 & 48 & 140 & 60\\
        
        \bottomrule
    \end{tabular}
    \caption{
    \textbf{Statistics of easy and hard queries for each LLM and dataset.}
    }
    \label{tb:query_difficulty}
\end{table*}

\section{Details of Datasets}\label{ap:dataset}

\paragraph{Trivia QA.} Trivia QA is a reading comprehension dataset, containing 650K question-answer-evidence triples. The questions and answers are gathered from trivia and quiz-league websites, and the evidences are collected from Web search results and Wikipedia articles. The collected questions are highly compositional and often require multi-hop reasoning, such as reasoning over time frames or making comparisons. In this work, we use it as a question-answering dataset and only utilize the questions as input.

\paragraph{NQ.} NQ is a question-answering dataset, containing 323K QA pairs. The questions are real user questions issued to Google search, and answers are found from Wikipedia by annotators. The dataset provides both long and short answers, where the long answers are spans extracted from Wikipedia articles and the short answers are entity or set of entities within the long answers. In this work, we consider short answers as ground truth.

\paragraph{Pop QA.} PopQA is a large-scale open-domain question answering dataset, consisting of 14k entity-centric QA pairs. The questions are created by sampling long-tail factual knowledge triples from Wikidata and converting them to natural language questions. 

\paragraph{Strategy QA.} Strategy QA is a question-answering dataset, containing 2.7K yes/no QA pairs. The questions are designed to be strategy questions, which are multi-step questions with implicit reasoning and a definitive answer.

\subsection{Statistics of Query Difficulty}\label{ap:query_difficulty}

For each dataset, we randomly sample 200 queries for all representation analyses. We split these 200 samples into easy and hard sets according to whether a model can correctly answer them without retrieval documents. Table~\ref{tb:query_difficulty} shows the statistics of the two sets of queries. Among them, NQ and Pop QA are considered harder than the others.

\section{Details of Retrieval Setting}\label{ap:retrieval}

In this work, we use MassiveDS as the retrieval database. MassiveDS is a vast, open-source database comprising approximately 1.4 trillion tokens. The dataset is built from a diverse collection of sources, including large-scale web crawl data as well as domain-specific corpora, to cover a wide variety of topics and writing styles. Each document in MassiveDS is a 256-word chunk of a passage, and was encoded by Contriever. For each query, we retrieve the top 20 documents using the retrieval and reranking pipeline introduced by \citet{NEURIPS2024_a5d8aba2}.

\section{Prompts Design}\label{ap:prompt}

We show the two prompts used in this paper below:

\begin{prompt}
    \textbf{Judging generated response.}

    \medskip

    You are an impartial evaluator tasked with judging the correctness of an answer. Your task: Determine if the model output is semantically and logically consistent with any of the ground truth answers.

\medskip
If it conveys the same meaning or correct information (even with different wording), mark it as correct.

\medskip
For yes/no questions, you only need to check whether the final [yes/no] prediction aligns with the ground truth or not.

\medskip
If the question is about the date, you need to check the consistency of the year, month, and day. It is OK if the model outputs the year only, but it is unacceptable if the generated one does not match the ground truth.

\medskip
If the model abstains from answering the question (e.g., saying the document does not contain sufficient information to answer the question), mark it as abstain.

\medskip
Respond ONLY with one of the following JSON objects:

\{"verdict": "correct"\}

\{"verdict": "incorrect"\}

\{"verdict": "abstain"\}
\end{prompt}

\begin{prompt}
\textbf{Classifying retrieved document.}
\medskip

    You are an objective evidence classifier. Given a user question, a list of possible answers, and a single document, decide whether the document is \textbf{relevant}, \textbf{distracting}, or \textbf{neutral} with respect to answering the query.

\medskip
\begin{itemize}[nosep]
    \item Do NOT produce a chain-of-thought. Provide only the required structured output (JSON, see schema below) and a concise 1-2 sentence rationale (no internal reasoning steps).
    \item Use external/world knowledge only to determine whether a document implicitly supports an answer via ordinary inference (e.g., a fact that implies resolvability, gender, date, etc.). Do not invent or hallucinate facts that are not in the document when justifying the label.
    \item Follow the definitions and heuristics below exactly.
\end{itemize}

\medskip
\#\# Required OUTPUT (JSON)

\medskip
Return a single JSON object with these fields only:

\medskip
```

\{

  "label": "relevant" | "distracting" | "neutral",

  "confidence": 0.00-1.00,
  
  "rationale": "\textlangle one- or two-sentence justification\textrangle",
  
  "supporting\_spans": ["\textlangle short excerpt(s) from the document that justify the judgment\textrangle"],
  
  "inference\_type": "direct" | "indirect" | "multi-hop" | "contradiction"

\}

```

\medskip
\begin{itemize}[nosep]
    \item label: one of relevant, distracting, neutral.
    \item confidence: numeric 0-1 reflecting how certain the label is (see scoring guidance below).
    \item rationale: no more than two sentences, explaining why the label was chosen.
    \item supporting\_spans: zero or more short text snippets taken verbatim from the document that most strongly support the label. If none, return [].
    \item inference\_type:
    \begin{itemize}[nosep]
        \item direct: the document explicitly states the answer.
        \item indirect: the document gives facts that strongly imply the answer.
        \item multi-hop: the document provides an intermediate hop (necessary fact) that, combined with other known facts, supports the answer.
        \item contradiction: the document asserts facts that contradict the correct answer.
    \end{itemize}
\end{itemize}
\medskip

\#\# Label definitions \& heuristics
\medskip

\#\#\# RELEVANT
\medskip

\begin{itemize}[nosep]
    \item The correct answer (or parts of answer) directly appeared in the document $\rightarrow$ set inference\_type = "direct".
    \item Or the document contains facts that clearly support the correct answer, either by single-step inference (set inference\_type = "indirect") or by providing a necessary intermediate hop for a multi-hop inference (set inference\_type = "multi-hop"). 
    \item If the doc contains intermediate facts that are required to get to the final answer (even though the final answer is not present), treat it as relevant (set inference\_type = "multi-hop").
    \item Provide supporting\_spans identifying the explicit sentence(s) or fact(s).
\end{itemize}

\medskip

\#\#\# DISTRACTING
\medskip

\begin{itemize}[nosep]
    \item The document asserts claims that would lead a reader away from the correct answer (\ie, it contradicts the correct answer or makes claims that support an incorrect candidate). Use inference\_type = "contradiction" if it explicitly contradicts.
    \item Or the document contains plausible but misleading facts that do not support the correct answer and could plausibly be mistaken for support (\eg, plausible but irrelevant facts presented as if they answer the question). Return supporting spans that illustrate the misleading claim.
    \item Or the document discusses other things that are related to some entities in the query, but does not provide hints for a reader to answer the question.
\end{itemize}

\medskip

\#\#\# NEUTRAL
\medskip

\begin{itemize}[nosep]
    \item The document is unrelated to the query.
\end{itemize}

\medskip

\#\# Confidence scoring guidance
\medskip

\begin{itemize}[nosep]
    \item $\geq 0.90$: explicit textual statement of the answer or a clear contradiction/distraction.
    \item $0.75 - 0.89$: strong indirect support or a strong but not explicit contradiction/distraction.
    \item $0.55 - 0.74$: moderate evidence (document gives facts that imply the answer but not overwhelmingly).
    \item $0.30 - 0.54$: weak or partial evidence, or small inconsistency; label should be conservative.
    \item $\leq 0.29$: little or no evidence; use for neutral decisions.
\end{itemize}

\medskip
Set a numeric value according to this guidance.
\end{prompt}

\subsection{Human Verification}

To validate the usage of LLM-as-a-Judge, we conduct human verification on LLMs' outputs. Specifically, for both response judging and retrieved document classification, we randomly select 50 data points and manually annotate them. We then compute the inter-annotator agreement between a human and LLMs. For response judging, we only observe 1 out of 50 with discrepancy between LLM and the human. For retrieved document classification, LLM achieves a 100\% agreement with human in the relevant set. And in the distracting set, we observe that only less than 5\% of distracting documents provide indirect information to ground answers. These results justify the data quality and validate the usage of LLM-as-a-Judge in data annotation.

\section{Example of Retrieved Document}\label{ap:examples}

We provide an example to demonstrate the differences between relevant, distracting, and random documents. In this example, the relevant document contains the exact ground truth answer, Jimi Hendrix, with sufficient context. In contrast, the distracting document does not provide any information about Jimi Hendrix, while the mentions of Monterey Pop festival and 1967 make the document have a high semantic similarity to the query.

\begin{prompt}

    \textbf{Query.} Who set fire to his guitar at the Monterey Pop festival in 1967?
    \medskip

    \textbf{Relevant.} Guitar showmanship to make a big thing of breaking the guitar. I bounced all over the stage with it and I threw the bits on the stage and I picked up my spare guitar and carried on as though I really had meant to do it." \textcolor{goodcolor}{Jimi Hendrix sometimes set fire to his guitar, most notably at the Monterey Pop Festival} when, apparently, he felt this was the only way he could upstage the destruction by Pete Townshend and Keith Moon of The Who during their set. On March 31, 1967 at performance at London Astoria Hendrix sustained hand burns and visited
    \medskip

    \textbf{Distracting.} doing them better. \textcolor{badcolor}{There's gonna be a showdown in California, at Monterey Pop, in July of 1967}, where we close out the show. Put on the earbuds, and take a ride on the Magic Bus. Turn it up, smash it up, burn it up! And thanks always for your comments, questions, and reviews! Note: we are using the American release dates and label imprint for all the songs on this episode's playlist.
    \medskip

    \textbf{Random.} 1. - Warren Beatty, Splendour in the Grass, 1961. Despite being championed by scenarist and playwright William Inge, Beatty nearly lost his first movie to the blonde cheeesecake. Troy had already taken Parrish from him. 2. - Richard Beymer, West Side Story, 1961. About as dumb an idea as Neil Diamond for Taxi Driver!
\end{prompt}

\begin{table*}[!t]
    \centering
    \small
    \begin{tabular}{lr@{}lr@{}lr@{}lr@{}l}
        \toprule
        LLM & \multicolumn{2}{c}{Trivia QA} & \multicolumn{2}{c}{NQ} & \multicolumn{2}{c}{Pop QA} & \multicolumn{2}{c}{Strategy QA}\\
        \midrule

        Gemma3-27B &  3.117 & *** & -1.787 & & -1.274 & & 14.375 & ***\\
        Llama4-17B & 5.361 & *** & -2.849 & & 2.135 & ** & 8.449 & ***\\
        Qwen3-80B & 9.525 & *** & 0.973 & & -0.663 & & 12.216 & ***\\
        
        \bottomrule
    \end{tabular}
    \caption{
    \textbf{Result of paired-sample one-tailed t-test.} We report the t-statistic, where its magnitude indicates how easy we can distinguish the two distributions. * $p<0.05$; ** $p<0.01$; *** $p<0.001$.
    }
    \label{tb:uncertainty}
\end{table*}

\begin{table*}[!t]
    \centering
    \small
    \begin{tabular}{lc@{\hskip6pt}cc@{\hskip6pt}cc@{\hskip6pt}cc@{\hskip6pt}c}
        \toprule
         & \multicolumn{2}{c}{Trivia QA} & \multicolumn{2}{c}{NQ} & \multicolumn{2}{c}{Pop QA} & \multicolumn{2}{c}{Strategy QA}\\
         \cmidrule(lr){2-3}
         \cmidrule(lr){4-5}
         \cmidrule(lr){6-7}
         \cmidrule(lr){8-9}
         LLM & Acc & AvgDist & Acc & AvgDist & Acc & AvgDist & Acc & AvgDist\\
        \midrule

        Gemma3-27B & 1.00 (0.00) & 0.78 (0.01) & 1.00 (0.00) & 0.77 (0.02) & 0.99 (0.01) & 0.68 (0.01) & 1.00 (0.00) & 0.73 (0.01)\\
        Llama4-17B & 1.00 (0.00) & 0.71 (0.01) & 1.00 (0.00) & 0.70 (0.01) & 1.00 (0.00) & 0.75 (0.01) & 0.99 (0.00) & 0.66 (0.01)\\
        Qwen3-80B & 0.98 (0.01) & 0.54 (0.02) & 0.97 (0.00) & 0.61 (0.01) & 1.00 (0.00) & 0.66 (0.02) & 0.97 (0.00) & 0.44 (0.01)\\
        
        \bottomrule
    \end{tabular}
    \caption{
    \textbf{In S1, the linear probe achieves both high accuracy and large distance.} 
    }
    \label{tb:h1}
\end{table*}
\begin{table*}[!t]
    \centering
    \small
    \begin{tabular}{lc@{\hskip6pt}cc@{\hskip6pt}cc@{\hskip6pt}cc@{\hskip6pt}c}
        \toprule
         & \multicolumn{2}{c}{Trivia QA} & \multicolumn{2}{c}{NQ} & \multicolumn{2}{c}{Pop QA} & \multicolumn{2}{c}{Strategy QA}\\
         \cmidrule(lr){2-3}
         \cmidrule(lr){4-5}
         \cmidrule(lr){6-7}
         \cmidrule(lr){8-9}
         LLM & Acc & AvgDist & Acc & AvgDist & Acc & AvgDist & Acc & AvgDist\\
        \midrule

        Gemma3-27B & 0.99 (0.00) & 0.31 (0.01) & 1.00 (0.00) & 0.37 (0.01) & 0.98 (0.01) & 0.41 (0.01) & 1.00 (0.00) & 0.40 (0.01)\\
        Llama4-17B & 1.00 (0.00) & 0.43 (0.01) & 0.95 (0.00) & 0.40 (0.00) & 1.00 (0.00) & 0.53 (0.02) & 1.00 (0.00) & 0.44 (0.01)\\
        Qwen3-80B & 0.74 (0.01) & 0.25 (0.01) & 0.83 (0.01) & 0.28 (0.00) & 0.89 (0.01) & 0.31 (0.00) & 0.97 (0.02) & 0.24 (0.01)\\
        
        \bottomrule
    \end{tabular}
    \caption{
    \textbf{In S2, the linear probe achieves a high accuracy but small distance.} 
    }
    \label{tb:h2}
\end{table*}

\section{Additional Experimental Results}\label{ap:result}

\subsection{Effect of Relevant Documents on Uncertainty}\label{ap:uncertainty}

We study how relevant documents affect the uncertainty of generated responses. In particular, we focus on queries that LLMs can answer correctly in both settings of without documents and with relevant documents.
We quantify uncertainty with length-normalized log-likelihood among generated tokens~\citep{guerreiro-etal-2023-looking}. Specifically, for a generation without retrieved documents, we compute the uncertainty score
\begin{align}
    s_\text{no\_doc}=\frac{1}{|\hat{y}|}\sum_{t=1}^{|\hat{y}|}\log(p_\theta(y_t|I,q,y_{<t})).
\end{align}
And for the generation with relevant documents, we compute 
\begin{align}
    s_\text{rel}=\frac{1}{|\hat{y}|}\sum_{t=1}^{|\hat{y}|}\log(p_\theta(y_t|I,q, S^{\text{rel}}_q,y_{<t})).
\end{align}

We hypothesize that the generations with relevant documents would have higher log-likelihood than generations without documents, \ie, $s_\text{rel}>s_\text{no\_doc}$. We perform paired-sample one-tailed t-test to verify this hypothesis. Table~\ref{tb:uncertainty} shows that t-test rejects the null hypothesis most of the time across datasets and models, suggesting that relevant documents help increase model confidence.

\subsection{Representations Separability Analyses}\label{ap:separability}

In Section~\ref{sec:result}, we primarily present our observations through PCA visualizations. To quantitatively substantiate these findings, we conduct linear probe experiments to measure the separability of representations induced by different context types. We report both linear probe accuracy and the average distance to the decision boundary as metrics of separability.

Let $\vw$ and $\vb$ denote the learned weight and bias of a linear probe, and let $\{\vx_i\}_{i=1}^N$ be the representations in the test set. We define the average distance to the decision boundary as
\begin{align}
d = \frac{1}{N} \sum_{i=1}^N \left| \frac{\vw^\top \vx_i + \vb}{\|\vw\|_2} \right|.
\end{align}

Based on the observations in Section~\ref{sec:result}, we evaluate the following settings:

\begin{enumerate}[label=\textbf{S\arabic*.}, itemsep=1pt]
    \item \textbf{No\_doc vs. Random}: High separability is expected (Observation~1).
    \item \textbf{No\_doc vs. Relevant}: Low separability is expected (Observation~2).
    \item \textbf{Relevant vs. Relevant + Distracting}: Low separability is expected (Observation~3).
    \item \textbf{Relevant vs. Random in early layer}: Low separability is expected (Observation~4).
    \item \textbf{No\_doc vs. Relevant in middle layer}: High separability is expected (Observation~5).
\end{enumerate}

\begin{table*}[!t]
    \centering
    \small
    \begin{tabular}{lc@{\hskip6pt}cc@{\hskip6pt}cc@{\hskip6pt}cc@{\hskip6pt}c}
        \toprule
         & \multicolumn{2}{c}{Trivia QA} & \multicolumn{2}{c}{NQ} & \multicolumn{2}{c}{Pop QA} & \multicolumn{2}{c}{Strategy QA}\\
         \cmidrule(lr){2-3}
         \cmidrule(lr){4-5}
         \cmidrule(lr){6-7}
         \cmidrule(lr){8-9}
         LLM & Acc & AvgDist & Acc & AvgDist & Acc & AvgDist & Acc & AvgDist\\
        \midrule

        Gemma3-27B & 0.66 (0.01) & 0.22 (0.01) & 0.62 (0.01) & 0.21 (0.01) & 0.65 (0.01) & 0.20 (0.01) & 0.98 (0.02) & 0.26 (0.00)\\
        Llama4-17B & 0.70 (0.01) & 0.20 (0.00) & 0.70 (0.01) & 0.24 (0.01) & 0.74 (0.03) & 0.21 (0.01) & 0.93 (0.01) & 0.18 (0.00)\\
        Qwen3-80B & 0.60 (0.01) & 0.26 (0.00) & 0.46 (0.02) & 0.21 (0.01) & 0.67 (0.01) & 0.27 (0.02) & 0.51 (0.01) & 0.21 (0.00)\\
        
        \bottomrule
    \end{tabular}
    \caption{
    \textbf{In S3, the linear probe achieves both low accuracy and small distance.} 
    }
    \label{tb:h3}
\end{table*}
\begin{table*}[!t]
    \centering
    \small
    \begin{tabular}{lc@{\hskip6pt}cc@{\hskip6pt}cc@{\hskip6pt}cc@{\hskip6pt}c}
        \toprule
         & \multicolumn{2}{c}{Trivia QA} & \multicolumn{2}{c}{NQ} & \multicolumn{2}{c}{Pop QA} & \multicolumn{2}{c}{Strategy QA}\\
         \cmidrule(lr){2-3}
         \cmidrule(lr){4-5}
         \cmidrule(lr){6-7}
         \cmidrule(lr){8-9}
         LLM & Acc & AvgDist & Acc & AvgDist & Acc & AvgDist & Acc & AvgDist\\
        \midrule

        Gemma3-27B & 0.49 (0.01) & 0.18 (0.01) & 0.52 (0.01) & 0.21 (0.00) & 0.63 (0.01) & 0.35 (0.01) & 0.50 (0.02) & 0.25 (0.01)\\
        Llama4-17B & 0.62 (0.01) & 0.13 (0.00) & 0.70 (0.01) & 0.14 (0.01) & 0.64 (0.01) & 0.24 (0.01) & 0.68 (0.01) & 0.16 (0.00)\\
        Qwen3-80B & 0.64 (0.02) & 0.14 (0.00) & 0.76 (0.02) & 0.18 (0.00) & 0.77 (0.02) & 0.25 (0.00) & 0.71 (0.01) & 0.21 (0.00)\\
        
        \bottomrule
    \end{tabular}
    \caption{
    \textbf{In S4, the linear probe achieves both low accuracy and small distance.} 
    }
    \label{tb:h4}
\end{table*}
\begin{table*}[!t]
    \centering
    \small
    \begin{tabular}{lc@{\hskip6pt}cc@{\hskip6pt}cc@{\hskip6pt}cc@{\hskip6pt}c}
        \toprule
         & \multicolumn{2}{c}{Trivia QA} & \multicolumn{2}{c}{NQ} & \multicolumn{2}{c}{Pop QA} & \multicolumn{2}{c}{Strategy QA}\\
         \cmidrule(lr){2-3}
         \cmidrule(lr){4-5}
         \cmidrule(lr){6-7}
         \cmidrule(lr){8-9}
         LLM & Acc & AvgDist & Acc & AvgDist & Acc & AvgDist & Acc & AvgDist\\
        \midrule

        Gemma3-27B & 1.00 (0.00) & 0.54 (0.01) & 1.00 (0.00) & 0.56 (0.00) & 1.00 (0.00) & 0.73 (0.00) & 1.00 (0.00) & 0.57 (0.00)\\
        Llama4-17B & 0.99 (0.01) & 0.57 (0.01) & 1.00 (0.00) & 0.46 (0.01) & 1.00 (0.00) & 0.61 (0.01) & 1.00 (0.00) & 0.46 (0.01)\\
        Qwen3-80B & 1.00 (0.00) & 0.44 (0.00) & 1.00 (0.00) & 0.48 (0.01) & 1.00 (0.00) & 0.59 (0.02) & 1.00 (0.00) & 0.50 (0.01)\\
        
        \bottomrule
    \end{tabular}
    \caption{
    \textbf{In S5, the linear probe achieves both high accuracy and large distance.} 
    }
    \label{tb:h5}
\end{table*}

For each setting, we split the data into training and test sets with an 80/20 ratio, apply PCA to reduce representations to 16 dimensions, and $\ell_2$-normalize each representation. We train a linear probe ten times using different random seeds and report the mean and standard deviation of each metric across runs.

Table~\ref{tb:h1} shows that in S1, the linear probe achieves both high accuracy and a large average distance, indicating strong separability between representations without documents and those with random documents, which supports Observation~1. In Table~\ref{tb:h2}, although the probe also attains high accuracy in S2, the average distance is substantially smaller than in S1. This suggests that representations without documents and with relevant documents are separable but remain close in the latent space, validating Observation~2. Tables~\ref{tb:h3} and~\ref{tb:h4} show that in both S3 and S4, the probe exhibits low accuracy and small average distance, indicating poor separability and supporting Observations~3 and~4. Finally, Table~\ref{tb:h5} shows that in S5, the probe again achieves high accuracy with a large average distance, confirming Observation~5 that in middle layers, representations without documents and with relevant documents are strongly separated.

\subsection{Additional Figures and Tables}\label{ap:additional_figure}

This section contains additional figures and tables that complement the results discussed in Section~\ref{sec:result}. Due to page limitations, these materials are presented here rather than in the main text. 

\begin{figure*}[!t]
    \centering
    \includegraphics[width=\textwidth]{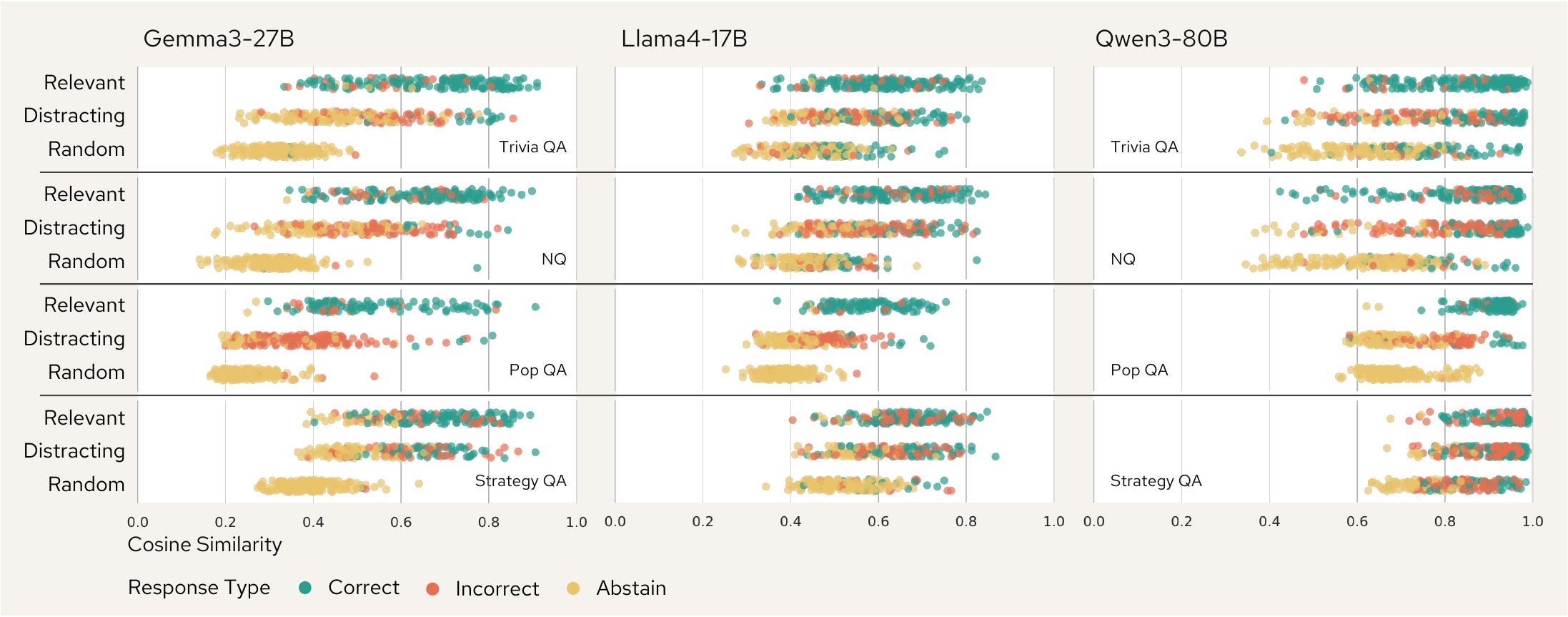} 
    \caption{\textbf{The complete result of Figure~\ref{fig:sim_vs_response}---relationship between cosine similarity and response type.} For each context type, we compute the cosine similarity between the representations of with-context prompts and query, and categorize responses as correct, incorrect, or abstain. The result shows that LLMs are more likely to abstain when context induces large representation shifts.}
    \label{fig:sim_vs_response_full}
\end{figure*}

\begin{figure*}[!t]
    \centering
    \includegraphics[width=\textwidth]{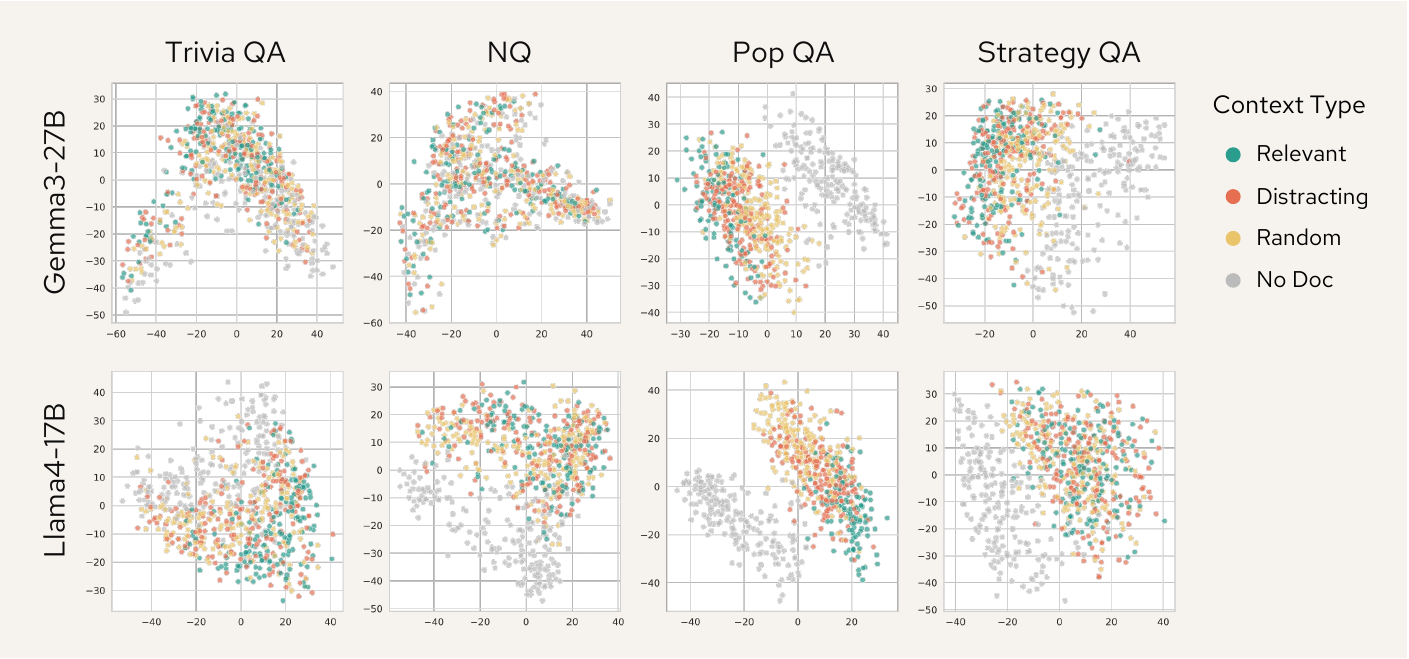} 
    \caption{\textbf{The complete result of Figure~\ref{fig:base_vs_it}---PCA of base models.} We apply PCA on representations of base models. The result shows that base LLMs do not have representation drift across different context types. Note that we only show the result of Gemma3-27B and Llama4-17B as Qwen3-80B did not release its base model.}
    \label{fig:base_vs_it_full}
\end{figure*}

\paragraph{Observation 1 \& 2.} See Figure~\ref{fig:sim_vs_response_full}, \ref{fig:base_vs_it_full}, and Table~\ref{tb:base_vs_it_full}.

\begin{figure*}[!t]
    \centering
    \includegraphics[width=\textwidth]{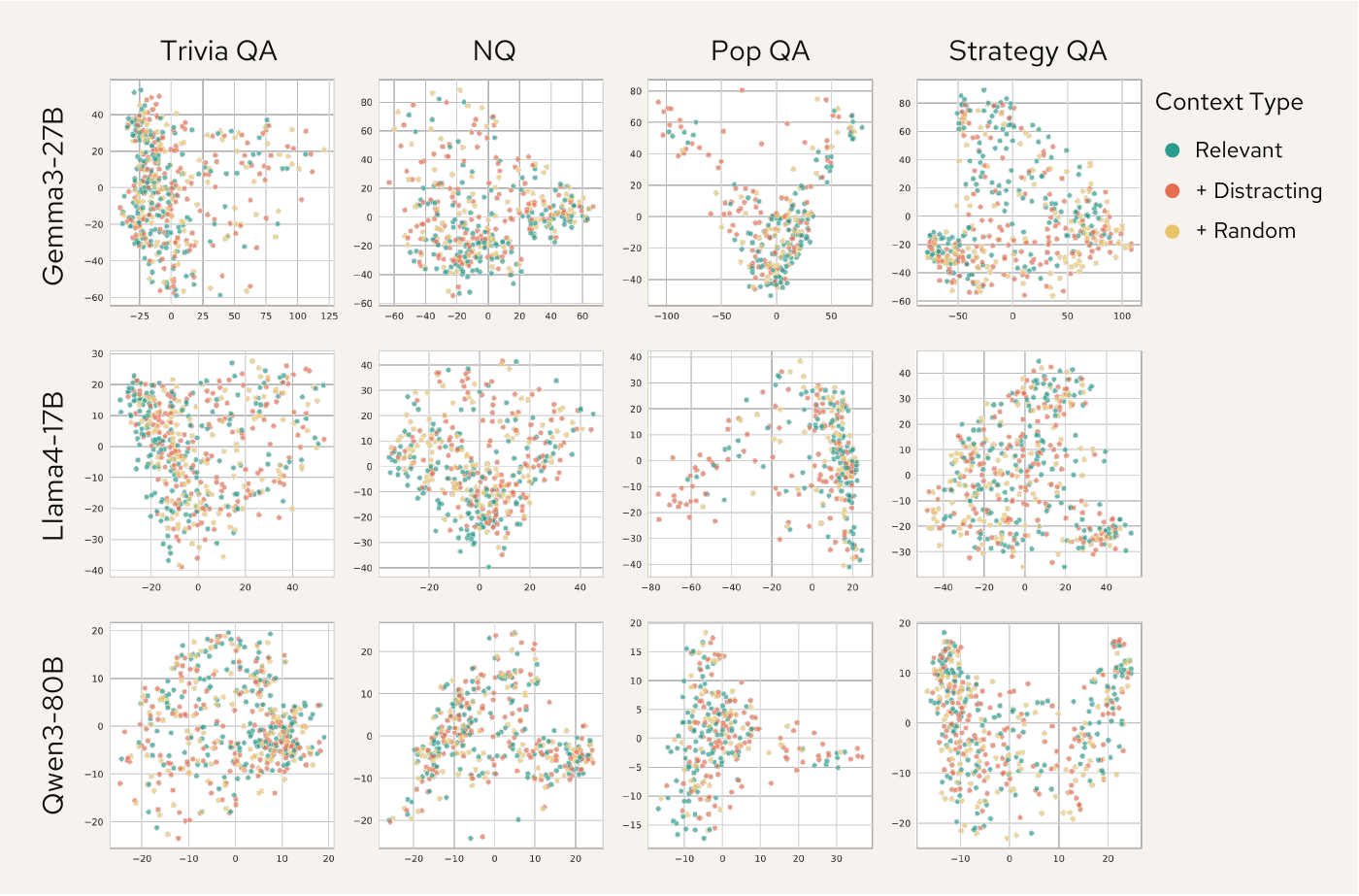} 
    \caption{\textbf{The complete result of Figure~\ref{fig:multiple_document}---PCA of multiple-document contexts.} We perform PCA on the last prompt token representations for multiple-document contexts that include one relevant document, and plot them alongside representations with only a relevant document in 2D. The result shows that representations remain similar when a relevant document is present, regardless of other context.}
    \label{fig:multiple_document_full}
\end{figure*}

\paragraph{Observation 3.} See Figure~\ref{fig:multiple_document_full}.

\begin{figure*}[!t]
    \centering
    \includegraphics[width=0.8\textwidth]{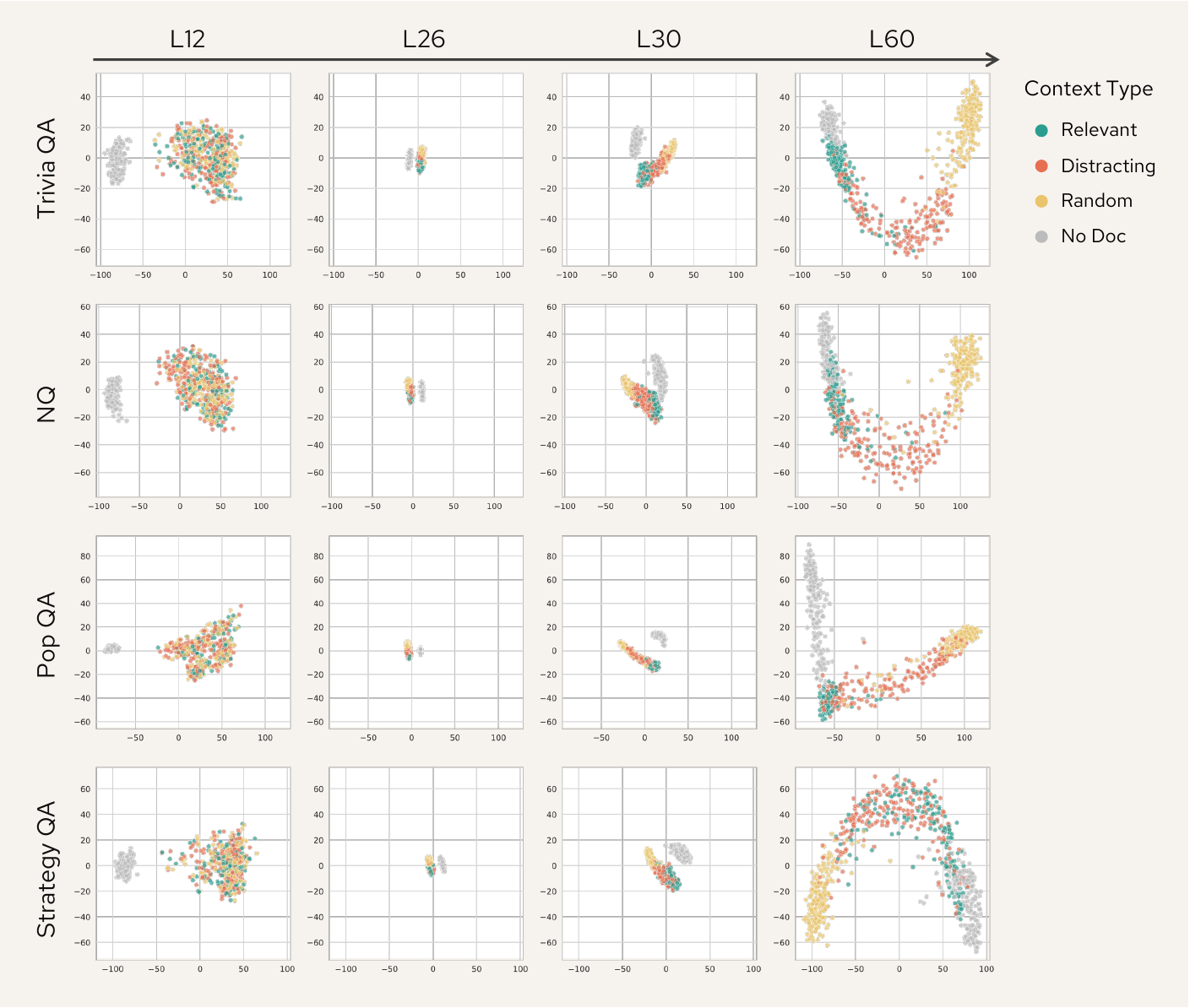} 
    \caption{\textbf{The complete result of Figure~\ref{fig:layer}---evolution of Gemma3-27B representations.} We perform PCA on the last prompt token representations of Gemma3-27B across different layers and plot them in 2D.}
    \label{fig:layer_gemma3_27B}
\end{figure*}

\begin{figure*}[!t]
    \centering
    \includegraphics[width=0.8\textwidth]{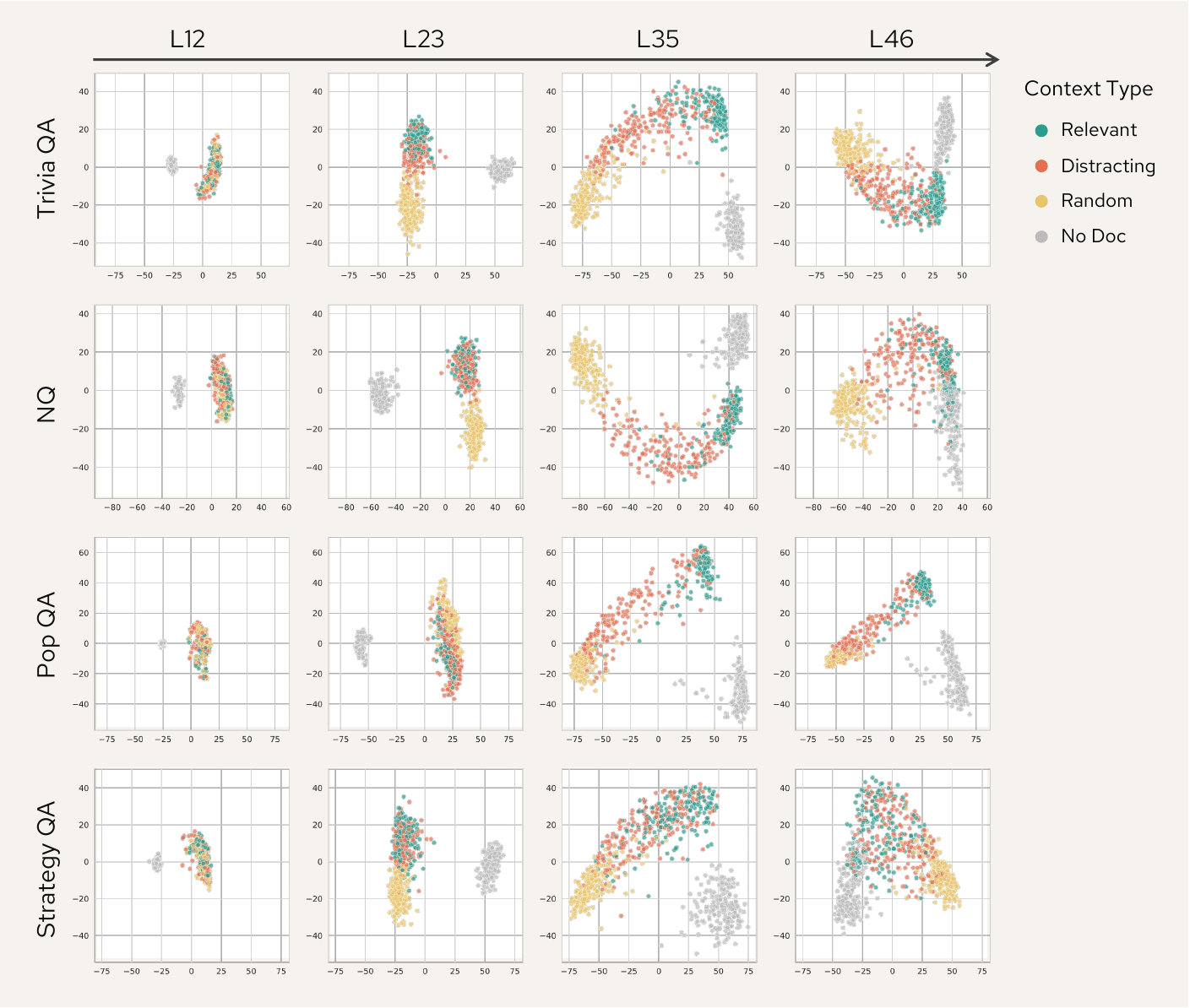} 
    \caption{\textbf{The complete result of Figure~\ref{fig:layer}---evolution of Llama4-17B representations.} We perform PCA on the last prompt token representations of Llama4-17B across different layers and plot them in 2D.}
    \label{fig:layer_llama4_17B}
\end{figure*}

\begin{figure*}[!t]
    \centering
    \includegraphics[width=0.8\textwidth]{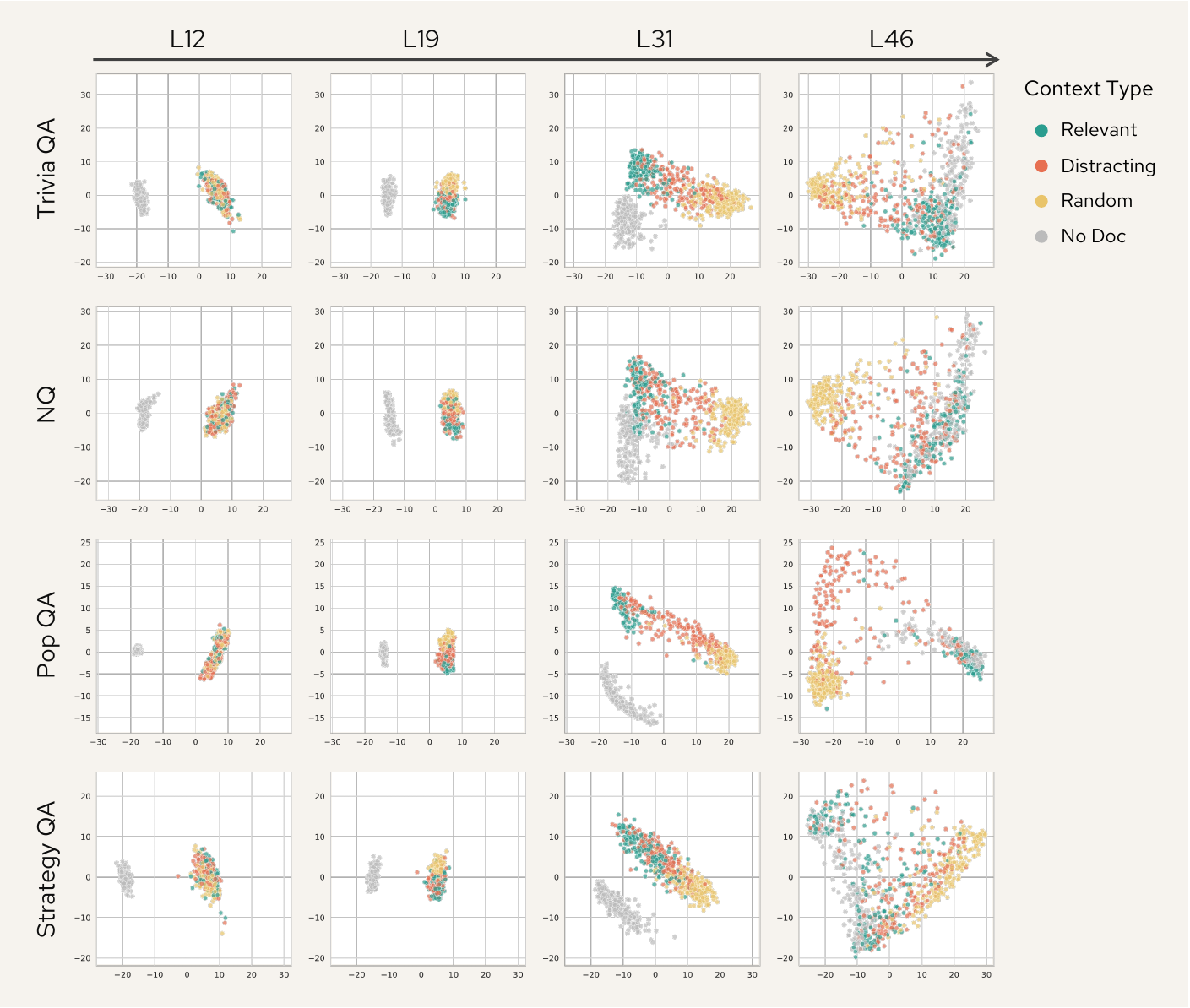} 
    \caption{\textbf{The complete result of Figure~\ref{fig:layer}---evolution of Qwen3-80B representations.} We perform PCA on the last prompt token representations of Qwen3-80B across different layers and plot them in 2D.}
    \label{fig:layer_qwen3_80B}
\end{figure*}

\begin{figure*}[!t]
    \centering
    \includegraphics[width=0.8\textwidth]{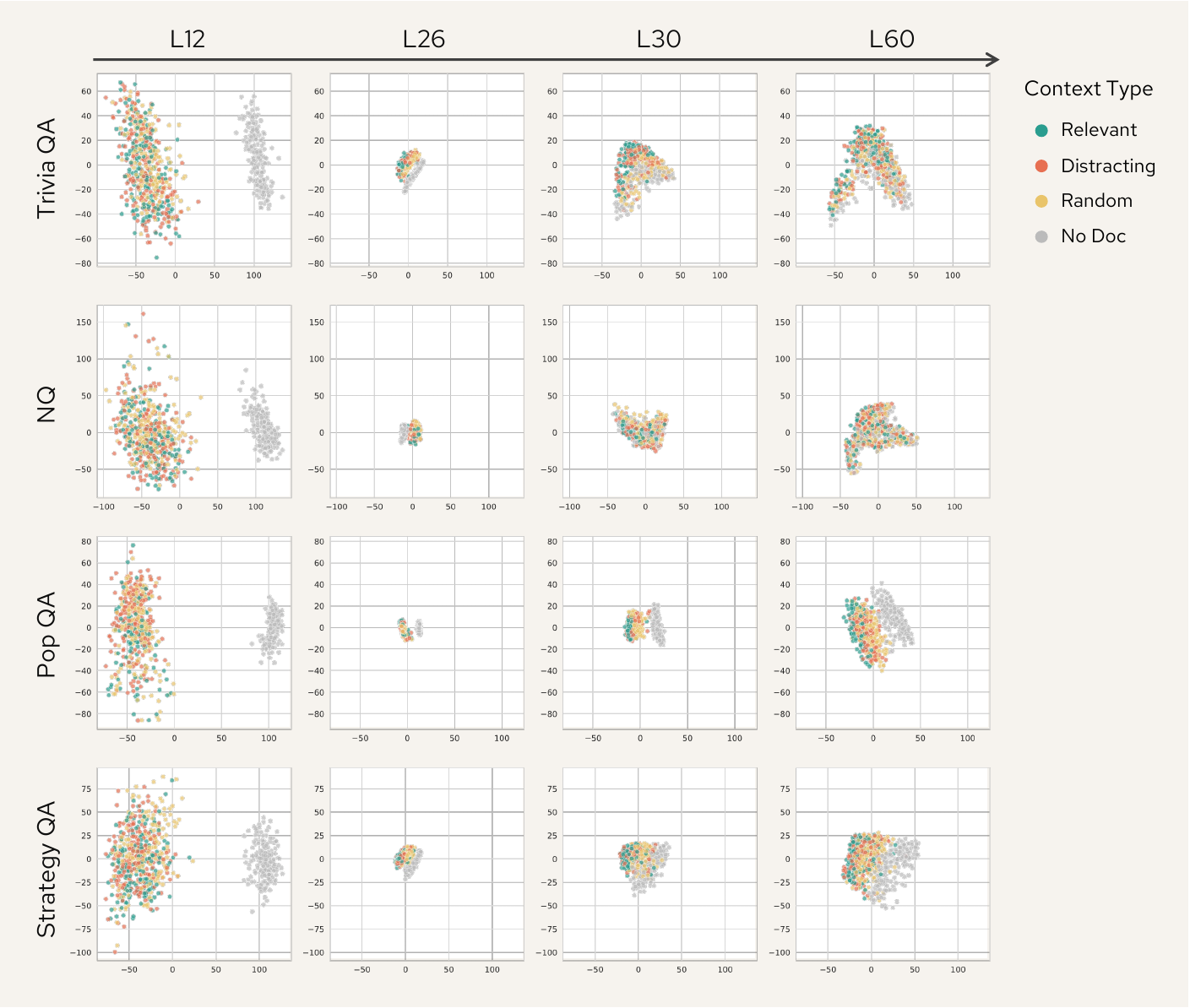} 
    \caption{\textbf{Evolution of Gemma3-27B-base representations.} We perform PCA on the last prompt token representations of Gemma3-27B-base across different layers and document types and plot them in 2D.}
    \label{fig:layer_gemma3_27B-base}
\end{figure*}

\begin{figure*}[!t]
    \centering
    \includegraphics[width=0.8\textwidth]{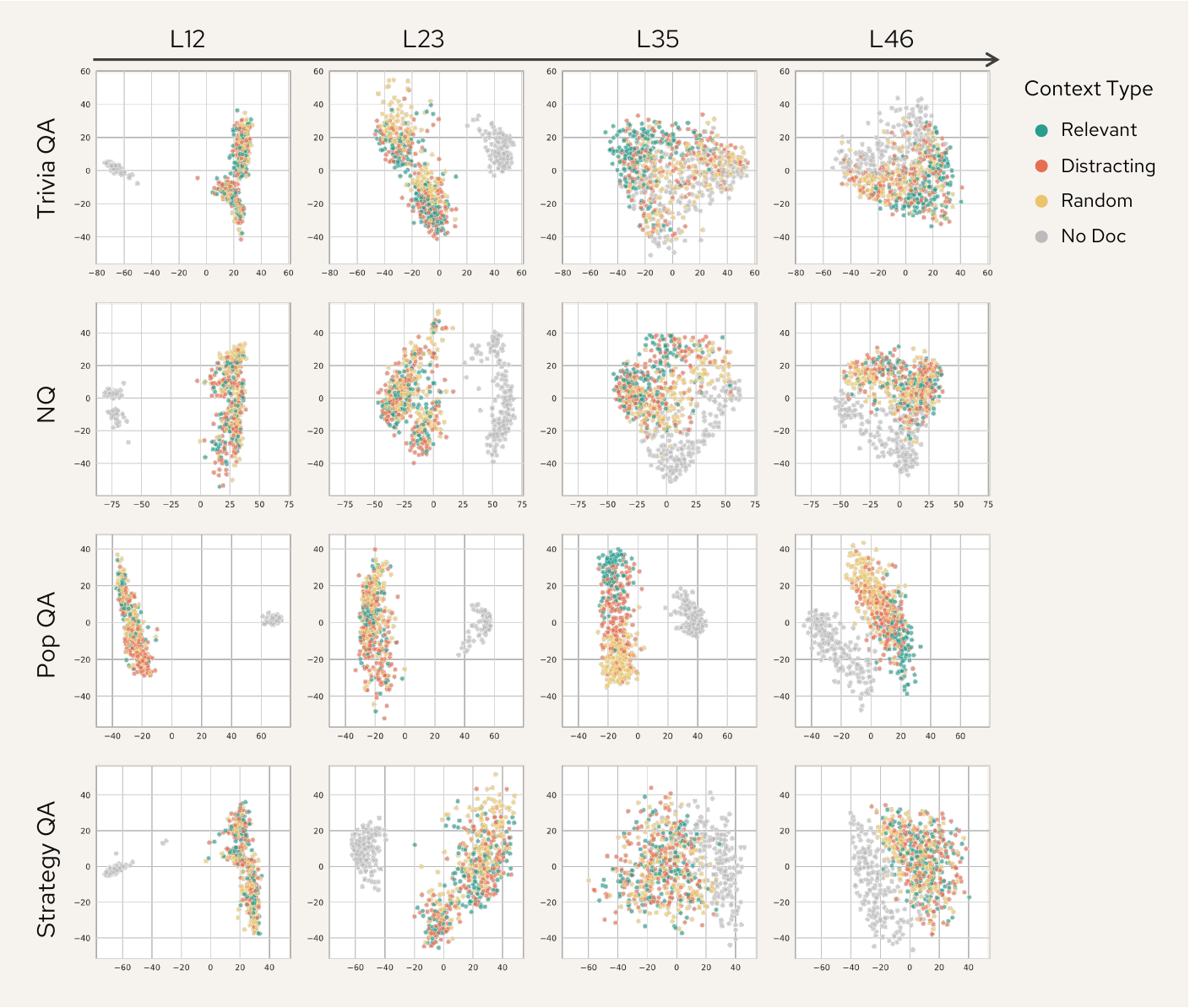} 
    \caption{\textbf{Evolution of Llama4-17B-base representations.} We perform PCA on the last prompt token representations of Llama4-17B-base across different layers and document types and plot them in 2D.}
    \label{fig:layer_llama4_17B-base}
\end{figure*}

\paragraph{Observation 4 \& 5.} See Figure~\ref{fig:layer_gemma3_27B} to \ref{fig:layer_llama4_17B-base}.

\begin{table*}[!t]
    \centering
    \small
    \resizebox{\textwidth}{!}{
    \begin{tabular}{l|cccccc|cccccc|cccccc|cccccc}
        \toprule
         & \multicolumn{6}{c|}{Trivia QA}  & \multicolumn{6}{c|}{NQ}  & \multicolumn{6}{c|}{Pop QA}  & \multicolumn{6}{c}{Strategy QA}\\
         & \multicolumn{3}{c}{Easy} & \multicolumn{3}{c|}{Hard} & \multicolumn{3}{c}{Easy} & \multicolumn{3}{c|}{Hard} & \multicolumn{3}{c}{Easy} & \multicolumn{3}{c|}{Hard} & \multicolumn{3}{c}{Easy} & \multicolumn{3}{c}{Hard}\\
         \cmidrule(lr){2-4}
         \cmidrule(lr){5-7}
         \cmidrule(lr){8-10}
         \cmidrule(lr){11-13}
         \cmidrule(lr){14-16}
         \cmidrule(lr){17-19}
         \cmidrule(lr){20-22}
         \cmidrule(lr){23-25}
         
         Context & Cor & Inc & Abs & Cor & Inc & Abs & Cor & Inc & Abs & Cor & Inc & Abs & Cor & Inc & Abs & Cor & Inc & Abs & Cor & Inc & Abs & Cor & Inc & Abs\\
        \midrule

        \rowcolor{goodcolor!30}\multicolumn{25}{c}{Gemma3-27B}\\
        \midrule
        \rowcolor{gray!15}\multicolumn{25}{c}{Base}\\
        \midrule
        Relevant & 92.4 & 4.2 & 3.4 & 52.5 & 35.6 & 11.9 & 81.9 & 10.2 & 7.9 & 43.9 & 39.5 & 16.6 & 92.4 & 2.0 & 5.6 & 76.7 & 20.4 & 2.9 & 74.5 & 16.4 & 9.1 & 44.6 & 42.2 & 13.2\\
        Distracting & 79.5 & 15.8 & 4.7 & 14.8 & 72.4 & 12.8 & 65.7 & 25.5 & 8.8 & 12.4 & 70.7 & 16.9 & 46.2 & 48.1 & 5.7 & 5.2 & 86.4 & 8.4 & 62.0 & 27.6 & 10.4 & 29.8 & 52.2 & 18.0\\
        Random & 89.4 & 6.8 & 3.8 & 15.6 & 65.4 & 19.0 & 66.5 & 12.2 & 21.3 & 11.1 & 53.1 & 35.8 & 71.5 & 24.2 & 4.3 & 9.7 & 85.4 & 4.9 & 65.6 & 20.9 & 13.5 & 23.4 & 55.8 & 20.8\\

        \midrule

        \rowcolor{gray!15}\multicolumn{25}{c}{Instruction-tuned}\\
        \midrule
        Relevant & 90.4 & 6.5 & 3.1 & 65.2 & 27.8 & 7.0 & 79.8 & 15.5 & 4.7 & 62.1 & 29.0 & 8.9 & 91.0 & 7.4 & 1.6 & 70.5 & 29.5 & 0 & 72.6 & 11.6 & 15.8 & 44.3 & 35.4 & 20.3\\
        Distracting & 8.5 & 29.7 & 61.8 & 0.7 & 25.1 & 74.2 & 8.0 & 39.0 & 53.0 & 0.6 & 34.3 & 65.1 & 3.6 & 52.7 & 43.7 & 0.6 & 61.6 & 37.8 & 22.7 & 19.7 & 57.6 & 8.6 & 23.3 & 68.1\\
        Random & 1.7 & 0.7 & 97.6 & 0 & 1.9 & 98.1 & 2.2 & 0.3 & 97.5 & 0.4 & 2.2 & 97.4 & 0.2 & 5.1 & 94.7 & 0.1 & 7.3 & 92.6 & 1.2 & 0.5 & 98.3 & 0 & 4.3 & 95.7\\

        \midrule
        \midrule

        \rowcolor{goodcolor!30}\multicolumn{25}{c}{Llama4-17B}\\
        \midrule
        \rowcolor{gray!15}\multicolumn{25}{c}{Base}\\
        \midrule
        Relevant & 93.5 & 6.2 & 0.3 & 67.2 & 32.2 & 0.6 & 84.5 & 15.4 & 0.1 & 58.7 & 40.9 & 0.4 & 95.4 & 4.6 & 0 & 79.6 & 20.4 & 0 & 89.1 & 10.9 & 0 & 34.1 & 65.5 & 0.34\\
        Distracting & 67.7 & 31.1 & 1.2 & 10.9 & 84.1 & 5.0 & 47.5 & 48.3 & 4.2 & 8.6 & 84.9 & 6.5 & 45.2 & 54.3 & 0.5 & 9.2 & 90.1 & 0.7 & 70.5 & 28.7 & 0.8 & 20.7 & 78.9 & 0.4\\
        Random & 91.0 & 8.0 & 1.0 & 7.6 & 86.4 & 6.0 & 76.4 & 19.5 & 4.1 & 15.6 & 73.0 & 11.4 & 74.0 & 25.5 & 0.5 & 20.8 & 78.7 & 0.5 & 74.1 & 24.9 & 1.0 & 20.8 & 78.6 & 0.6\\

        \midrule

        \rowcolor{gray!15}\multicolumn{25}{c}{Instruction-tuned}\\
        \midrule
        Relevant & 89.7 & 8.5 & 1.8 & 62.8 & 30.3 & 6.9 & 85.9 & 12.8 & 1.3 & 64.0 & 30.6 & 5.4 & 92.1 & 6.5 & 1.4 & 79.8 & 15.5 & 4.7 & 76.3 & 17.1 & 6.6 & 35.3 & 56.2 & 8.5\\
        Distracting & 34.6 & 26.4 & 39.0 & 11.4 & 37.0 & 51.6 & 33.5 & 34.5 & 32.0 & 5.1 & 44.3 & 50.6 & 2.5 & 26.1 & 71.4 & 0.5 & 28.0 & 71.5 & 38.5 & 24.8 & 36.7 & 13.2 & 45.6 & 41.2\\
        Random & 38.7 & 2.5 & 58.8 & 8.7 & 12.1 & 79.2 & 34.5 & 5.3 & 60.2 & 4.0 & 15.5 & 80.5 & 0.4 & 0.8 & 98.8 & 0 & 0.9 & 99.1 & 13.8 & 6.1 & 80.1 & 8.2 & 10.9 & 80.9\\

        \bottomrule
    \end{tabular}
    }
    \caption{
    \textbf{The complete result of Table~\ref{tb:base_vs_it}} We report the percentage of correct (Cor), incorrect (Inc), and abstain (Abs) responses for both base and instruction-tuned LLMs. The result shows that instruction-tuned LLMs tend to abstain when the retrieval document is distracting or random, even if they can answer with the query alone. Note that we only show the result of Gemma3-27B and Llama4-17B as Qwen3-80B did not release its base model.
    }
    \label{tb:base_vs_it_full}
\end{table*}


\end{document}